\documentclass{article}

\usepackage[nonatbib,final]{neurips_2021}

\usepackage[utf8]{inputenc}
\usepackage[T1]{fontenc}
\usepackage{booktabs}
\usepackage{amsmath,amsthm, bm}
\usepackage{amssymb, amsfonts}
\usepackage{nicefrac,microtype}
\usepackage{tikz,cancel}
\usepackage{wasysym,graphicx}
\usepackage{float,mathtools}
\usepackage{commath, mathrsfs}
\usepackage{enumitem}
\usepackage{physics, braket}
\usepackage{subfigure}
\usepackage{dcolumn}
\usepackage[ruled,vlined]{algorithm2e}
\usepackage{xcolor}
\usepackage{hyperref,url}
\hypersetup{colorlinks=true, allcolors=blue}

\theoremstyle{remark}
\newtheorem{remark}{Remark}

\renewcommand{\figref}[2][]{{{\figurename}}\,\ref{#2}#1}
\renewcommand{\theequation}{{\arabic{equation}}}
\renewcommand{\eqref}[1]{Eq.(\ref{#1})} 

\theoremstyle{plain}

\newtheorem{prop}{Proposition}

\DeclareMathOperator*{\argmin}{arg\,min}

\def\diag{{\rm diag}}

\def\bPsi{\mathbf{\Psi}}
\def\bPhi{\mathbf{\Phi}}

\def\bSigma{{\bm \Sigma}}

\def\bLambda{{\bm \Lambda}}

\def\bmu{{\bm \mu}}

\def\0{{\bf 0}}
\def\1{{\bf 1}}

\def\a{{\bf a}}
\def\b{{\bf b}}

\def\v{\mathbf v}
\def\u{\mathbf u}
\def\w{{\mathbf w}}
\def\x{{\mathbf x}}

\def\A{{\bf A}}

\def\C{{\bf C}}

\def\G{{\bf G}}

\def\I{{\bf I}}

\def\M{{\bf M}}
\def\Q{{\bf Q}}

\def\R{{\bf R}}

\def\W{{\mathbf W}}
\def\X{{\mathbf X}}

\def\U{\mathbf U}
\def\bxi{{\bm \xi}}
\def\bchi{{\bm \chi}}

\title{Out-of-Distribution Generalization in Kernel Regression}

\author{
  Abdulkadir Canatar \\
  Department of Physics\\
  Harvard University\\
  Cambridge, MA 02138 \\
  \texttt{canatara@g.harvard.edu} \\
  \And
  Blake Bordelon \\
  John A. Paulson School of Engineering and Applied Sciences \\
  Harvard University \\
  Cambridge, MA 02138 \\
  \texttt{blake\_bordelon@g.harvard.edu} \\
  \And
  Cengiz Pehlevan \\
  John A. Paulson School of Engineering and Applied Sciences \\
  Harvard University \\
  Cambridge, MA 02138 \\
  \texttt{cpehlevan@g.harvard.edu}
}

\begin{document}

\maketitle

\begin{abstract}

In real word applications, the data generating process for training a machine learning model often differs from what the model encounters in the test stage. Understanding how and whether machine learning models generalize under such distributional shifts remains a theoretical challenge. Here, we study generalization in kernel regression when the training and test distributions are different using the replica method from statistical physics. We derive an analytical formula for the out-of-distribution  generalization error applicable to any kernel and real datasets. We identify an overlap matrix that quantifies the mismatch between distributions for a given kernel as a key determinant of generalization performance under distribution shift. Using our analytical expressions we elucidate various generalization phenomena including possible improvement in generalization when there is a mismatch. We develop procedures for optimizing training and test distributions for a given data budget to find best and worst case generalizations under the shift.  We present applications of our theory to real and synthetic datasets and for many kernels. We compare results of our theory applied to Neural Tangent Kernel with simulations of wide networks and show agreement. We analyze linear regression in further depth.
\end{abstract}

\section{Introduction}

Machine learning models are trained to accurately predict on previously unseen samples. A central goal of machine learning theory has been to understand this generalization performance \cite{Vapnik1998,mohri2018foundations}. While most of the theory in this domain focused on generalizing in-distribution, i.e. when the training examples are sampled from the same distribution as that of the test examples, in real world applications there is often a mismatch between the training and the test distributions \cite{amodei2016concrete}. Such difference, even when small, may lead to large effects in generalization performance \cite{gonzalez2015mismatched,su2019one,engstrom2019exploring,recht2019imagenet,hendrycks2019benchmarking}. 

In this paper, we provide a theory of out-of-distribution (OOD)  generalization for kernel regression \cite{wahba1990spline,evgeniou2000regularization,shawe2004kernel} applicable to any kernel and real datasets. Besides being a popular machine learning method itself, kernel regression is recovered from an infinite width limit of neural networks \cite{jacot2018neural}, making our results relevant to understanding OOD generalization in deep neural networks. Indeed, it has been argued that understanding generalization in kernel methods is necessary for understanding generalization in deep learning \cite{belkin2018understand}.

Using methods from statistical physics \cite{mezard1987spin} and building on techniques developed in a recent line of work \cite{bordelon2020spectrum,canatar2020spectral}, we obtain an analytical expression for generalization in kernel regression when test data is drawn from a different distribution than the training distribution. Our results apply to average or typical case learning curves and describe numerical experiments well including those on real data. In contrast, most previous works focus on worst-case OOD generalization bounds in the spirit of statistical learning theory \cite{quinonero2009dataset,ben2010theory,tsipras2018robustness,krueger2020outofdistribution,arjovsky2019invariant,arjovsky2020out,redko2019advances}. 


  

We show examples of how mismatched training and test distributions affect generalization and in particular, demonstrate that the mismatch may improve test performance. We use our analytical expression to develop a procedure for minimizing generalization error with respect to the training/test distributions and present applications of it. We study OOD generalization with a linear kernel in detail and present examples of how dimensional reduction of the training distribution can be helpful in generalization, including shifting of a double-descent peak \cite{belkin2019reconciling}. We present further analyses of various models in Supplementary Information (SI). 

\section{OOD Generalization Error for Kernel Regression from the Replica Method}

We consider kernel regression in the setting where training examples are sampled i.i.d. from a distribution $p(\x)$ but the test examples are drawn from a different distribution $\tilde p(\x)$. We derive our main analytical formula for the generalization error here, and illustrate and discuss its implications in the following sections.


\subsection{Problem setup}\label{Psetup}

We consider a training set $\mathcal{D} = \{(\x^\mu, y^\mu)\}_{\mu = 1}^P$ of size $P$ where the $D$-dimensional inputs $\x \in \mathbb{R}^{D}$ are drawn independently from a training distribution $p(\x)$ and  the noisy labels are generated from a target function $y^\mu = \bar f(\x^\mu) + \epsilon^\mu$ where the noise covariance is $\mathbb{E}[\epsilon^\mu\epsilon^\nu] = \varepsilon^2\delta_{\mu\nu}$.  Kernel regression model is trained through minimizing the training error:
\begin{align}
    f^*_\mathcal{D} = \argmin_{f \in \mathcal{H}} \frac{1}{2} \sum_{\mu = 1}^P \big(f(\x^\mu) - y^\mu\big)^2 + \lambda \braket{f,f}_\mathcal{H},
\end{align}
where {subscript $\mathcal{D}$ emphasizes the dataset dependence.} Here $\mathcal{H}$ is a Reproducing Kernel Hilbert Space (RKHS) associated with a positive semi-definite kernel $K(\x,\x'): \mathbb{R}^{D}\times \mathbb{R}^{D} \to \mathbb{R}$, and  $\braket{.,.}_\mathcal{H}$ is the Hilbert inner product.

The generalization error on the test distribution $\tilde p(\x)$ is given by $E_g(\mathcal{D}) = \int d\x \, \tilde p(\x) \left(f^*_\mathcal{D}(\x) - {\bar f(\x)} \right)^2$. We note that this quantity is a random variable whose value depends on the sampled training dataset. We calculate its average over the distribution of all datasets with sample size $P$:
\begin{align}
    E_g = \mathbb{E}_\mathcal{D}\bigg[\int d\x \, \tilde p(\x) \left(f^*_\mathcal{D}(\x) - {\bar f(\x)} \right)^2\bigg].
\end{align}
As $P$ grows, we expect fluctuations around this average to fall and $E_g(\mathcal{D})$ to concentrate around $E_g$. We will demonstrate this concentration in simulations.

\subsection{Overview of the calculation}

We calculate $E_g$ using the replica method from statistical physics of disordered systems \cite{mezard1987spin,engel2001statistical}. Details of calculations are presented in Section \ref{SI:calculation}. Here we give a short overview. Readers may choose to skip this part and proceed to the main result in Section \ref{main}.

Our goal is to calculate the dataset averaged estimator $f^*(\x) \equiv \mathbb{E}_\mathcal{D}{f_\mathcal{D}^*(\x)}$ and its covariance $\mathbb{E}_\mathcal{D}\big[{\big(f_\mathcal{D}^*(\x)-f^*(\x)\big)\big(f_\mathcal{D}^*(\x')-f^*(\x')\big)}\big]$. From these quantities, we can reconstruct $E_g$ using the bias-variance decomposition of $E_g = B+V$, where  $B = \int d\x\, \tilde p(\x) \left(f^*(\x) - \bar f(\x) \right)^2$ and $V = \mathbb{E}_\mathcal{D}\big[\int d\x \, \tilde p(\x) \left(f_\mathcal{D}^*(\x) - f^*(\x) \right)^2\big]$. 

For this purpose, it will be convenient to work with a basis in the RKHS defined by Mercer's theorem \cite{RasmussenWilliams}. One can find a (possibly infinite dimensional) complete orthonormal basis $\{\phi_\rho\}_{\rho = 1}^M$ for $L^2(\mathbb{R}^D)$ with respect to the training probability distribution $p(\x)$ due to Mercer's theorem \cite{RasmussenWilliams} such that
\begin{align}
    K(\x,\x') = \bPhi(\x)^\top\bLambda\bPhi(\x'),\quad \int\,d\x\, p(\x) \bPhi(\x)\bPhi(\x)^\top = \I,
\end{align}
where we defined $M\times M$ diagonal eigenvalue matrix $\bLambda_{\rho\gamma} = \eta_\rho\delta_{\rho\gamma}$ and the column vector $\bPhi(\x) = \begin{pmatrix}\phi_1(\x),\phi_2(\x),\dots,\phi_M(\x)\end{pmatrix}$. Eigenvalues and eigenfunctions satisfy the integral eigenvalue equation \begin{align}\label{eq:eigenvalue}\int\,d\x' p(\x') K(\x,\x')\phi_\rho(\x') = \eta_\rho \phi_\rho(\x).\end{align} We note that the kernel might not express all eigenfunctions if the corresponding eigenvalues vanish. Here we assume that all kernel eigenvalues are non-zero for presentation purposes, however in Section \ref{SI:calculation} we consider the full case. We also define a feature map via the column vector $\bPsi_\rho(\x) \equiv (\bLambda^{1/2}\bPhi(\x))_\rho = \sqrt{\eta_\rho}\phi_\rho(\x)$ so that $\braket{\bPsi(\x),\bPsi(\x)}_\mathcal{H} = \I$. The complete basis $\bPhi(\x)$ can be used to decompose the target function:
\begin{align}\label{eq:target}
    \bar f(\x) = \bar\a^\top \bPhi(\x) = \bar \w ^\top\bPsi(\x),
\end{align}
where $\bar w =(\bLambda^{-1/2}\bar\a)$, and $\bar \a$ and $\bar \w$ are vectors of coefficients. A function belongs to the RKHS $\mathcal{H}$ if it has finite Hilbert norm $\braket{f,f}_\mathcal{H}^2 = \a^\top\bLambda^{-1}\a < \infty$. 

With this setup, denoting the estimator as $f(\x) =  \w^\top \bPsi(\x)$ and the target function $\bar f(\x)$ as given in \eqref{eq:target}, kernel regression problem reduces to minimization of the energy function $H(\w;\mathcal{D}) \equiv \frac{1}{2\lambda} \sum_{\mu=1}^P\left((\bar \w -\w) ^\top\bPsi(\x^\mu)+\epsilon^\mu\right)^2 + \frac{1}{2}\Vert\w\Vert^2_2.$ with optimal estimator weights $\w^*_\mathcal{D} = \argmin_{\w} H(\w;\mathcal{D})$. 
We again emphasize the dependence of the optimal estimator weights $\w^*_{\mathcal D}$ to the particular choice of training data $\mathcal{D}$. 

We map this problem to statistical mechanics by defining a Gibbs distribution $\propto e^{-\beta H(\w;\mathcal{D})}$ over estimator weights $\w$ which concentrates around the kernel regression solution $\w^*_\mathcal{D}$ as $\beta\to\infty$. This can be used to calculate any function  $O(\w^*;\mathcal{D})$ by the following trick:
\begin{align}
    O(\w^*;\mathcal{D}) = \lim_{\beta\to\infty}\frac{\partial}{\partial J} \log Z[J;\beta, \mathcal{D}]\big|_{J=0} ,\quad Z[J;\beta, \mathcal{D}] = \int d\w e^{-\beta H(\w;\mathcal{D}) + J O(\w)},
\end{align} 
{where $Z$ is the normalizer of the Gibbs distribution, also known as the partition function. Next, we want to compute the average of $\mathbb{E}_\mathcal{D}O(\w^*;\mathcal{D})$ which requires computing $\mathbb{E}_\mathcal{D}\log Z$. Further, experience from the study of physics of disordered systems suggests that the logarithm of the partition function concentrates around its mean (is self-averaging) for large $P$  \cite{mezard1987spin}, making our theory applicable to the typical case. However, this average is analytically hard to calculate due to the partition function appearing inside the logarithm. To proceed, we resort to the replica method from statistical physics \cite{mezard1987spin}, which uses the equality $\mathbb{E}_\mathcal{D}\log Z  = \lim_{n\rightarrow 0}\frac{\mathbb{E}_\mathcal{D}{Z^n}-1}{n}$. The method proceeds by calculating the right hand side for integer $n$, analytically continuing the resulting expression to real valued $n$, and performing the limit. While non-rigorous, the replica method has been successfully used in the study of the physics of disordered systems \cite{mezard1987spin} and machine learning theory \cite{advani2013statistical}. A crucial step in our computation is approximating $\w^\top\bPsi(\x)$
 as a Gaussian random variable via its first and second moments when averaged over the training distribution $p(\x)$. It has been shown that this approximation yields perfect agreement with experiments \cite{sompolinsky1999statistical,bordelon2020spectrum,canatar2020spectral}. Details of our calculation is given in Supplementary Section \ref{SI:calculation}. }
 


\subsection{Main Result}\label{main}

The outcome of our statistical mechanics calculation (Section \ref{SI:calculation}) which constitutes our main theoretical result is presented in the following proposition.

\begin{prop}\label{prop:main}
Consider the kernel regression problem outlined in Section \ref{Psetup}, where the model is trained on $p(\x)$ and tested on $\tilde p(\x)$. Consider the Mercer decomposition of the RKHS kernel $ K(\x,\x') = \bPhi(\x)^\top\bLambda\bPhi(\x')$, where we defined $M\times M$ ($M$ possibly infinite) diagonal eigenvalue matrix $\bLambda_{\rho\gamma} = \eta_\rho\delta_{\rho\gamma}$ and the column vector $\bPhi(\x) = \begin{pmatrix}\phi_1(\x),\phi_2(\x),\dots,\phi_M(\x)\end{pmatrix}$, with $\int\,d\x\, p(\x) \bPhi(\x)\bPhi(\x)^\top = \I$. Also consider an expansion of the target function in the Mercer basis $\bar f(\x) = \bar\a^\top \bPhi(\x)$.

The dataset averaged out-of-distribution generalization error is given by: 
\begin{gather}\label{eq:main_gen_error}
    E_g = E_g^{0,p(\x)} + \frac{\gamma' - \gamma}{1-\gamma}\varepsilon^2 +  \kappa^2\bar\a^\top (P\bLambda+\kappa\I)^{-1} \mathscr{O}'(P\bLambda+\kappa\I)^{-1}\bar\a,\nonumber\\
    \kappa = \lambda  + \kappa\Tr\,(P+\kappa\bLambda^{-1})^{-1},\quad \gamma = P\Tr\, (P+\kappa\bLambda^{-1})^{-2},\quad \gamma' = P\Tr \mathscr{O}(P+\kappa\bLambda^{-1})^{-2},
\end{gather}
where $\kappa$ must be solved self-consistently, and we defined the $M\times M$ overlap matrix
\begin{align}
    \mathscr{O}_{\rho\gamma} = \int\,d\x\, \tilde p(\x) \phi_\rho(\x)\phi_\gamma(\x),\quad \mathscr{O}' = \mathscr{O} - \frac{1-\gamma'}{1-\gamma}\I.
\end{align}
Here $E_g^{0,p(\x)}$ denotes the generalization error when both training and test distributions are matched to $p(\x)$ and is given by:
\begin{align}
    E_g^{0,p(\x)} = \frac{\gamma}{1-\gamma}\varepsilon^2 + \frac{\kappa^2}{1-\gamma}\bar\a^\top (P\bLambda+\kappa\I)^{-2}\bar\a,
\end{align}
which coincides with the findings of \cite{bordelon2020spectrum,canatar2020spectral}. Further, the expected estimator is:
\begin{align}
{f^*(\x;P)}= \sum_\rho \frac{P\eta_\rho}{P\eta_\rho + \kappa}\bar a_\rho \phi_\rho(\x).
\end{align}
\end{prop}

Several remarks are in order.
\begin{remark}
Our result is general in that it applies to any kernel, data distribution and target function. When applied to kernels arising from the infinite width limit of neural networks \cite{jacot2018neural}, the information
about the architecture of the network is in the kernel spectrum $\bLambda$ and the target weights $\bar\a$ obtained by projecting the target function onto kernel eigenbasis.
\end{remark}

\begin{remark}
Formally, the replica computation requires a thermodynamic limit in which $P\to\infty$, where variations in $E_g$ due to the sampling of the training set become negligible.
The precise nature of the limit depends on the kernel and the data
distribution, and includes scaling of other variables such as $D$ and $M$ with $P$. We will give examples of such limits in SI. However, we observe in simulations that our formula predicts average learning curves accurately for even as low as a few samples.
\end{remark}

\begin{remark}
We recover the result obtained in \cite{bordelon2020spectrum,canatar2020spectral} when the training and test distributions are the same ($\mathscr{O} = \I$ which implies $E_g = E_g^{0,p(\x)}$). 
\end{remark}

\begin{remark}
{\it Mismatched training and test distributions may improve test error.} Central to our analysis is the shifted overlap matrix $\mathscr{O}'$ which may have negative eigenvalues and hence cause better generalization performance when compared to in-distribution generalization.
\end{remark}

\begin{remark}
The estimator $f^*$ only depends on the training distribution and our theory predicts that the estimator eventually approaches to the target $f^* \to \bar f$ for large enough $P$ which performs perfectly on the training distribution unless there are out-of-RKHS components in target function. The latter case is studied in depth in Section \ref{SI:calculation}.
\end{remark}

\begin{remark}
While stated for a scalar output, our result can be trivially generalized to a vector output by simply adding the error due to each component of the output vector, as described in \cite{bordelon2020spectrum}.
\end{remark}

 Next we  analyze this result by studying various examples.

\section{Applications to Real Datasets}\label{sec:real_datasets}

First, we test our theory on real datasets and demonstrate its broad applicability. To do that, we define Dirac training and test probability measures {for a fixed dataset $\mathcal{D}$ of size $M$} with some point mass on each of the data points {$p(\mathbf{x}) =  \sum_{\x^\mu \in \mathcal{D}} p^{\mu}\delta(\mathbf{x}-\mathbf{x}^{\mu})$ and $\tilde p(\mathbf{x}) =  \sum_{\x^\mu \in \mathcal{D}} \tilde p^{\mu}\delta(\mathbf{x}-\mathbf{x}^{\mu})$}. With this measure, the kernel eigenvalue problem in \eqref{eq:eigenvalue} becomes
$\bm K \text{diag}(\bm p) \bm\Phi = \bm\Phi \bm\Lambda$, where $\text{diag}({\bm p})$ denotes the diagonal matrix of probability masses of each data point. {In this setup the number of eigenfunctions, or eigenvectors, are equal to the size of entire dataset $\mathcal{D}$.} Once the eigenvalues $\mathbf{\Lambda}$ and the eigenvectors $\bm{\Phi}$ have been identified, we 
compute the target function coefficients by projecting the target data $\mathbf{y}_c$ onto these eigenvectors $\mathbf{\overline{a}_c} = \bm{\Phi}^{\top} \text{diag}(\bm p) \mathbf{y}_c$ for each target $c=1,\ldots,C$. Once all of these ingredients are obtained, theoretical learning curves can be computed using  Proposition \ref{prop:main}. This procedure is similar to the one used in \cite{bordelon2020spectrum,canatar2020spectral}.

\subsection{Shift in test distribution may help or hurt generalization}

\begin{figure}[t]
    \centering
    \subfigure[Optimized Test Measures]{\includegraphics[width=0.31\linewidth]{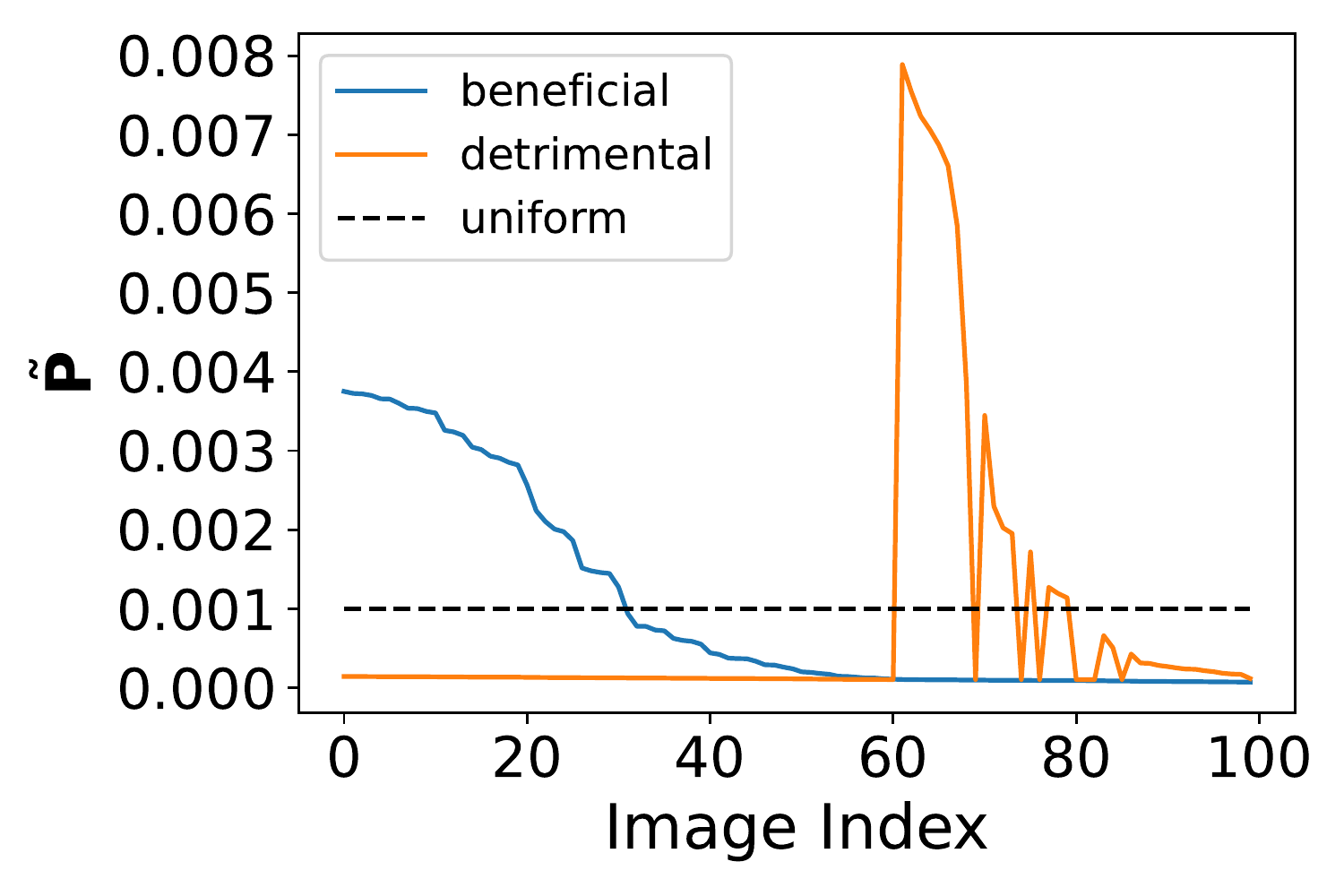}}
    \subfigure[Measure change matrix $\mathscr{O}'$]{\includegraphics[width=0.68\linewidth]{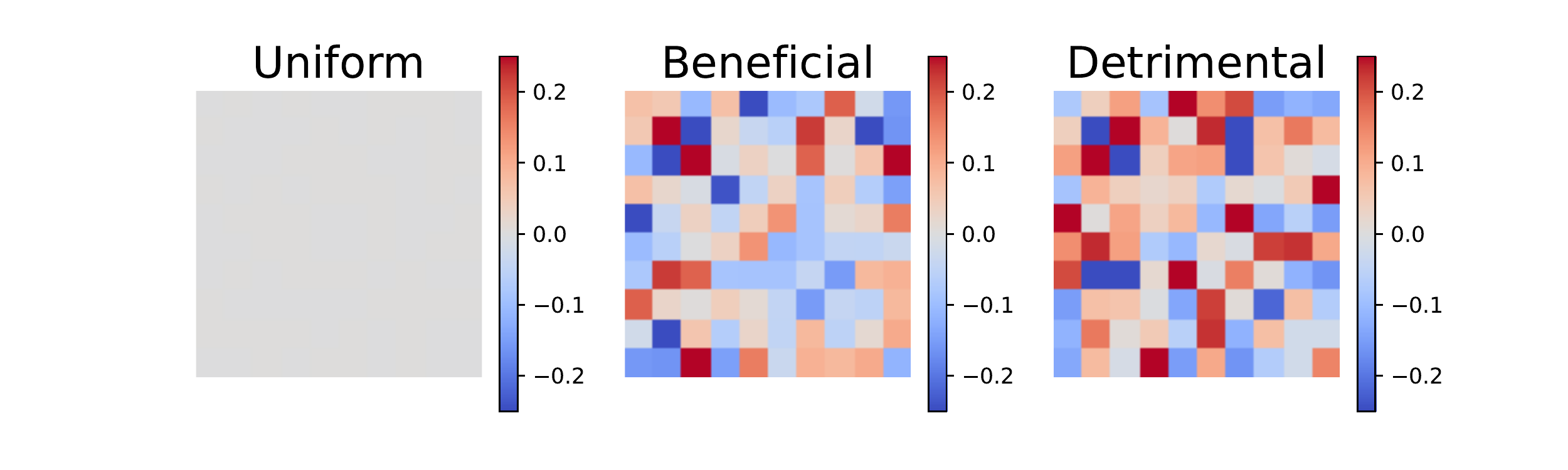}}
    \subfigure[Spectrum of $\mathscr{O}'$]{\includegraphics[width=0.32\linewidth]{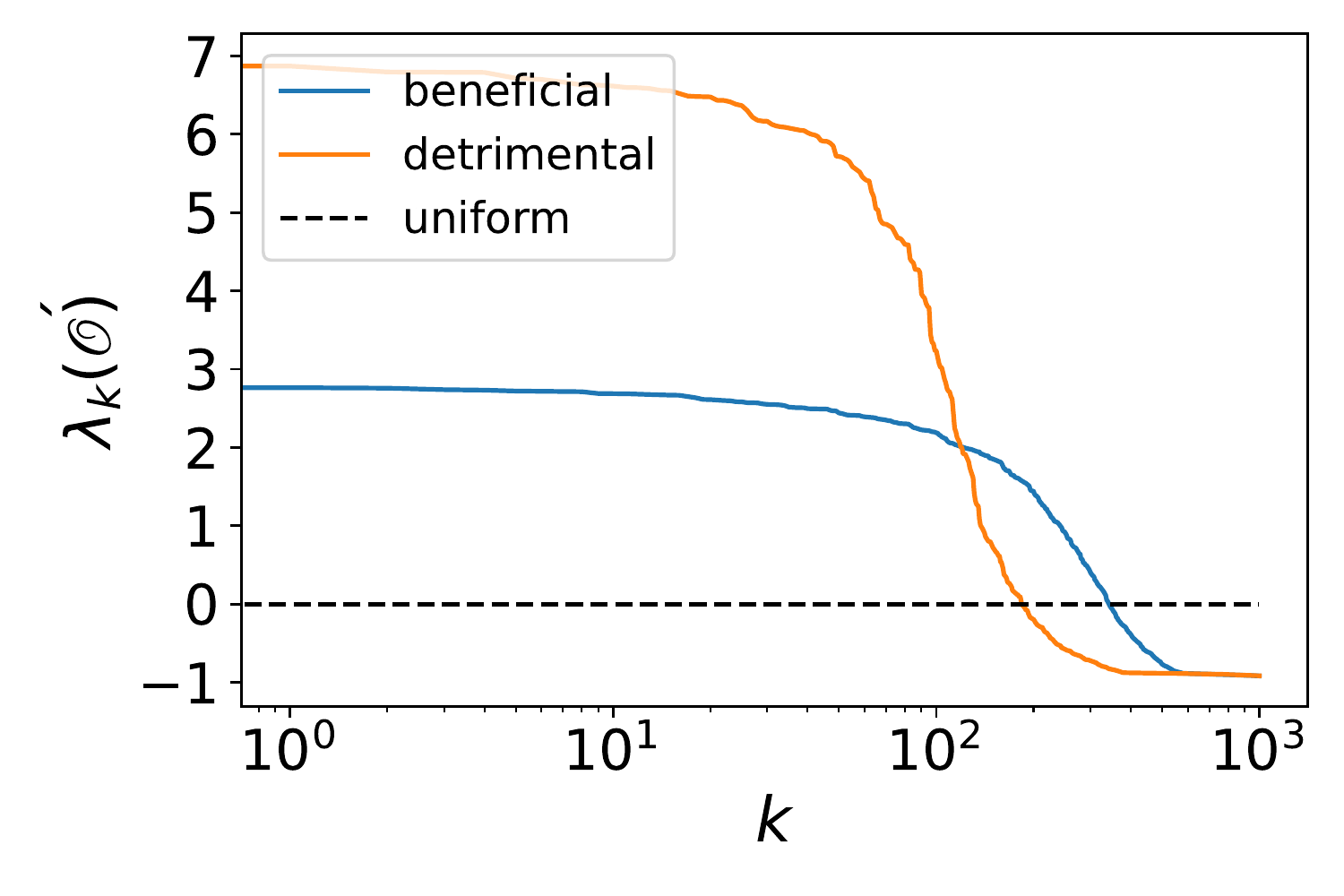}}
    \subfigure[Learning Curves]{\includegraphics[width=0.32\linewidth]{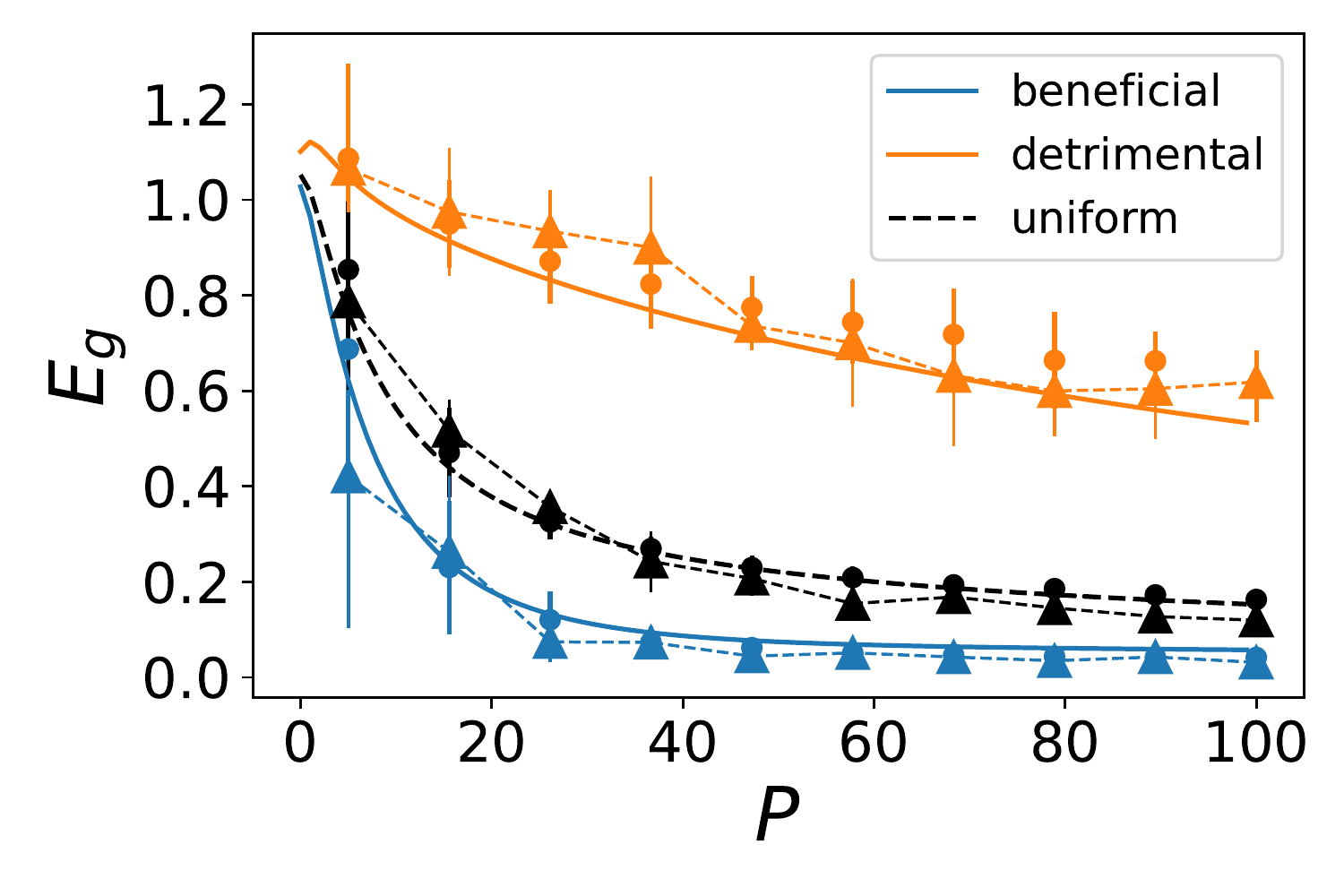}}
    \caption{Shifts in the test distribution can be understood through the matrix $\mathscr{O}'$. (a) We ran gradient descent (beneficial) and gradient ascent ({detrimental}) on the theoretical generalization error $E_g$ with respect to the test measure on $1000$ MNIST images. (a) The final probability of the first $100$ images sorted by probability mass for the beneficial measure.  (b) For these three distributions (uniform, beneficial, {detrimental}), the measure change matrices $\mathscr{O}'$ are plotted for the top $10$ modes. The beneficial and {detrimental} matrices are roughly opposite. (c) The spectrum of $\mathscr{O}'$ reveals both negative and positive eigenvalues which demonstrates that increase or a decrease in the generalization error is possible. (d) The learning curves show the predicted generalization error (lines) along with {trained neural network (triangles) and corresponding NTK regression experiments (dots)}. Error bars indicate standard deviation over $35$ random trials. }
    \label{fig:opt_test}
\end{figure}

To analyze the influence of distribution shift on generalization error, we first study the problem of a fixed training distribution and a shifting test distribution. While one may naively expect that a shift in the test distribution would always result in worse performance, we demonstrate that mismatch can be either beneficial or detrimental to generalization. A similar point was made in \cite{gonzalez2015mismatched}. 

Since our generalization error formula is a differentiable function with respect to the training measure, 
we numerically optimize $E_g$ with respect to $\{ \tilde p^{\mu} \}$ using gradient descent to find beneficial test measures where the error is smaller than for the case where training and test measures are matched. Similarly, we perform gradient ascent to find {detrimental} test measures, which give higher test error than the matched case. 

The result of this procedure is shown in \figref{fig:opt_test}. We perform gradient descent (labeled beneficial) and gradient ascent (labeled detrimental) on $M = 1000$ MNIST digits $8$'s and $9$'s, where target outputs are binary $\{-1,1\}$ corresponding to two digits. We use a depth $3$ ReLU fully-connected neural network with $2000$ hidden units at each layer and its associated Neural Tangent Kernel (NTK) which are computed using the NeuralTangents API \cite{neuraltangents2020}.
The probability mass assigned to the first $50$ points are provided in \figref{fig:opt_test}(a). We see that points given high mass for the beneficial test distribution are given low probability mass for the {detrimental} test distribution and vice versa. We plot the measure change matrix $\mathscr{O}'$ which quantifies the change in the generalization error due to distribution shift. Recall that in the absence of noise, $E_g$ is of the form $E_g = E_g^{\text{matched}} + \bm v^\top \mathscr{O}' \bm v$ for a vector $\bm v$ which depends on the task, kernel, training measure and sample size $P$. Depending on the eigenvalues and eigenvectors of $\mathscr{O}'$, and the vector $\bm v$, the quadratic form $\bm v^\top \mathscr{O}' \bm v$ may be positive or negative, resulting in improvement or detriment in the generalization error. In \figref{fig:opt_test}(b), we see that the $\mathscr{O}'$ matrix for the beneficial test measure appears roughly opposite that of the {detrimental} measure. This is intuitive since it would imply the change in the generalization error to be opposite in sign for beneficial and {detrimental} measures $\bm v^\top \left( - \mathscr{O}' \right) \bm v = - \bm v^\top \mathscr{O}' \bm v$. The spectrum of the matrix, shown in \figref{fig:opt_test}(c), reveals that it possesses both positive and negative eigenvalues. We plot the theoretical and experimental learning curves in \figref{fig:opt_test}(d). As promised, the beneficial (blue) measure has lower generalization error while the {detrimental} measure (orange) has higher generalization error than the matched case (black). Experiments show excellent match to our theory. {In Section \ref{SI:further_real_data}, we present another application of this procedure to adversarial attacks during testing.
}

\subsection{Optimized Training Measure for Digit Classification with Wide Neural Networks}

\begin{algorithm}[t]\label{alg1}
\SetAlgoLined
\KwResult{GET\_LOSS($\bm z \in \mathbb{R}^M$, $\bm K \in \mathbb{R}^{M\times M}$, $\bm y \in \mathbb{R}^{M\times C}$, $\lambda \in \mathbb{R}_+$) }
\Indp Compute normalized distribution $\bm{p} = \text{softmax}(\bm z)$\;
  Diagonalize on Train Measure $\bm K \text{diag}(\bm p) \bm\Phi = \bm\Phi \bm\Lambda$ with $\bm\Phi^\top \text{diag}(\bm p) \bm\Phi = \bm I$ \;
   Get Target Function Weights $\mathbf{\overline{a}} = \bm\Phi^\top \text{diag}(\bm p) \bm y$\;
   Get Overlap Matrix $\mathscr{O} = \frac{1}{M} \bm\Phi^\top \bm\Phi$ \;
   Solve Implicit Equation \ $\kappa =$ ODE-INT$\left[ \dot{\kappa} = \lambda + \kappa\sum_{k} \frac{\Lambda_{kk}}{\Lambda_{kk} P + \kappa} - \kappa  \right]$ \;
   \textbf{return} $E_g(\kappa, P, \bm\Lambda, \mathscr{O})$ (see Proposition \ref{eq:main_gen_error})\;
\Indm
\;
\KwResult{OPT\_MEASURE($\bm K \in \mathbb{R}^{M \times M}$, $\bm y \in \mathbb{R}^M$, $P$, $T$ ,$\eta$, $\lambda$) }
\Indp
 Initialize $\bm z = \bm 0 \in \mathbb{R}^{M}$ , $t = 0$\;
 Diagonalize Kernel on Uniform Measure $\frac{1}{M} \bm K = \tilde{\bm\Phi} \tilde{\bm\Lambda} \tilde{\bm\Phi}^\top$\;
 \While{$t < T$}{
    $\bm z = \bm z - \eta \ \text{GRAD}[\text{GET\_LOSS}(\bm z, \bm K, \bm y, \lambda)]$\;
    $t = t + 1$\;
 }
 \textbf{return} $\text{softmax}(\bm z)$ \;
 \caption{Optimizing Training Measure at sample size $P$}
\end{algorithm}

Often in real life tasks test distribution is fixed while one can alter the training distribution for more efficient learning. Therefore, we next study how different training distributions affect the generalization performance of kernel regression when the test distribution is fixed. We provide pseudocode in Algorithm \ref{alg1} which describes a procedure to optimize the expected generalization error with respect to a training distribution of $P$ data points. This optimal training distribution has been named the dual distribution to $\tilde p(x)$ \cite{gonzalez2015mismatched}. All of the operations in the computation of $E_g$ support automatic differentiation with respect to logits $\bm z \in \mathbb{R}^M$ that define the training distribution through $\text{softmax}(\bm z)$, allowing us to perform gradient descent on the training measure \cite{jax2018github}.

\begin{figure}[t]
    \centering
    \subfigure[Optimized Measure]{\includegraphics[width=0.32\linewidth]{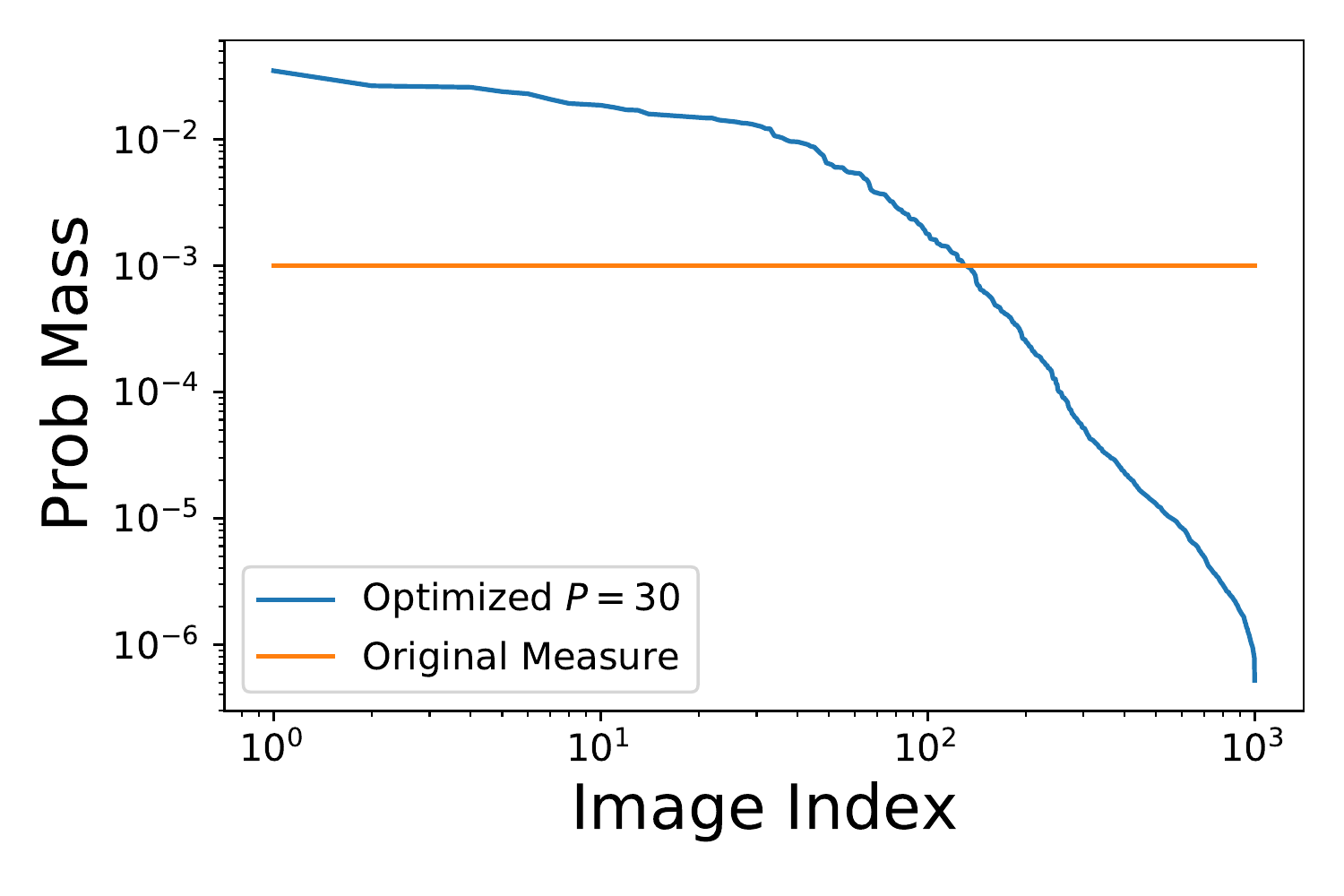}}
    \subfigure[High Mass Training Digits]{\includegraphics[width=0.32\linewidth]{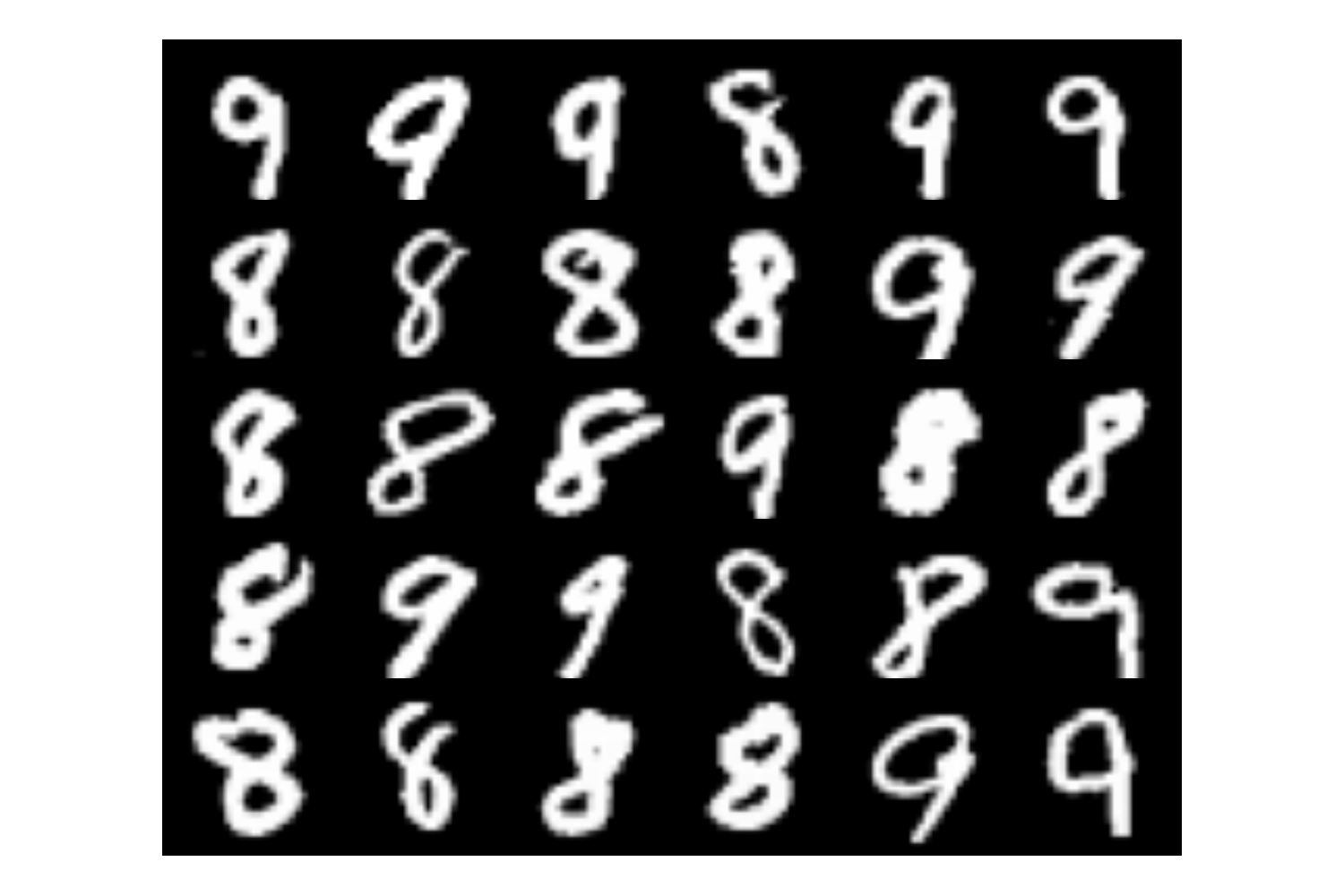}}
    \subfigure[Low Mass Training Digits]{\includegraphics[width=0.32\linewidth]{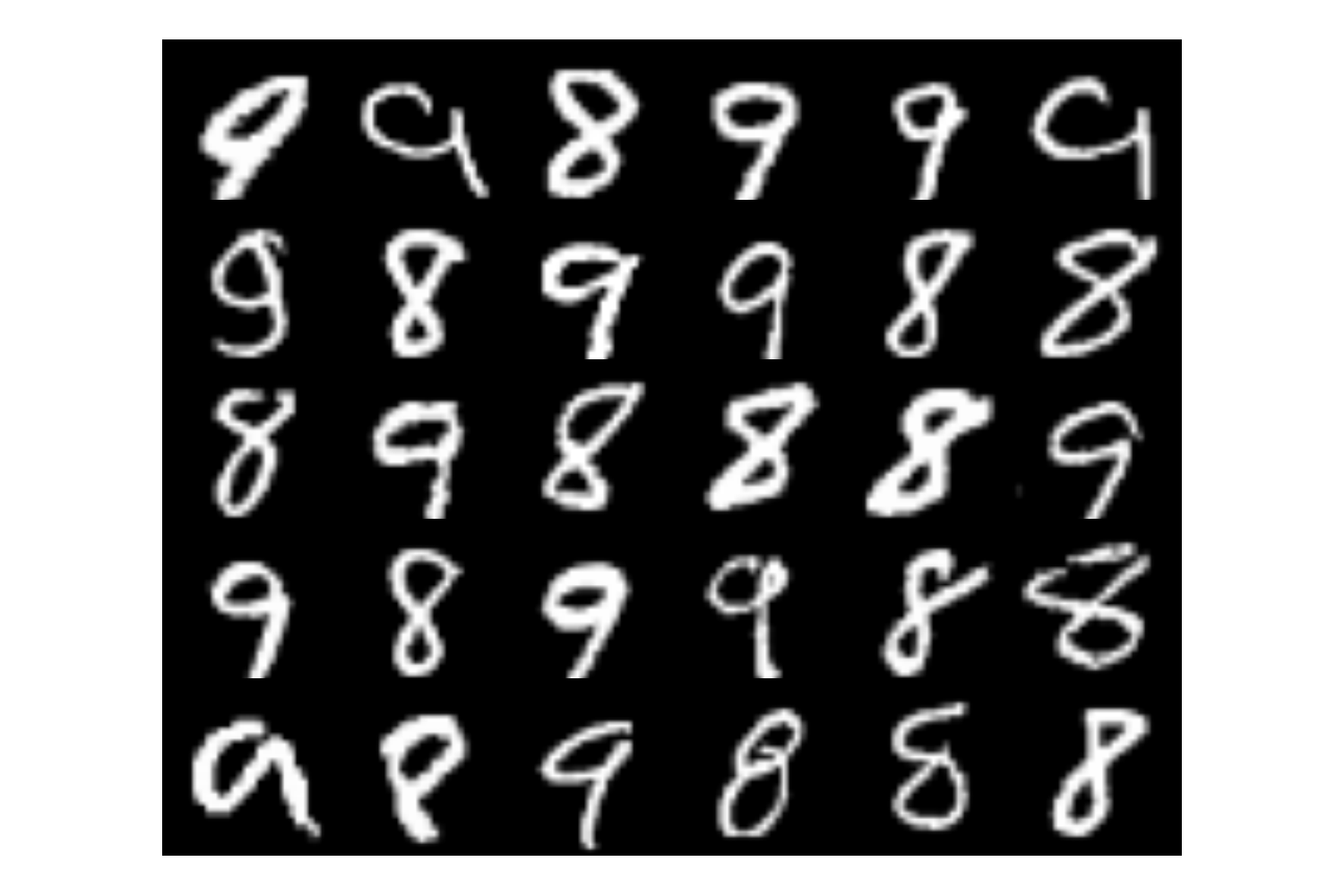}}
    \subfigure[Learning Curves]{\includegraphics[width=0.32\linewidth]{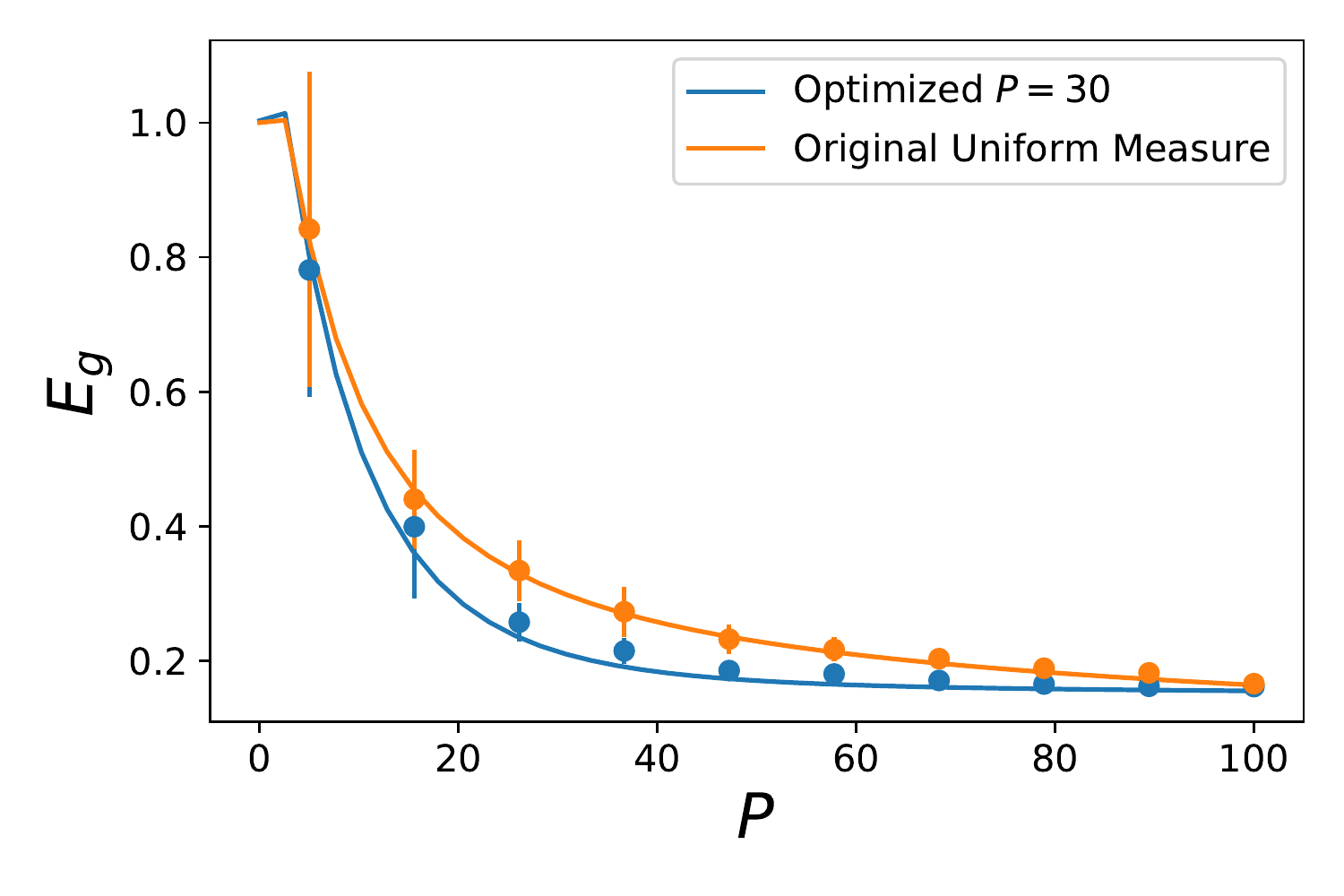}}
    \subfigure[Within Class Distance]{\includegraphics[width=0.32\linewidth]{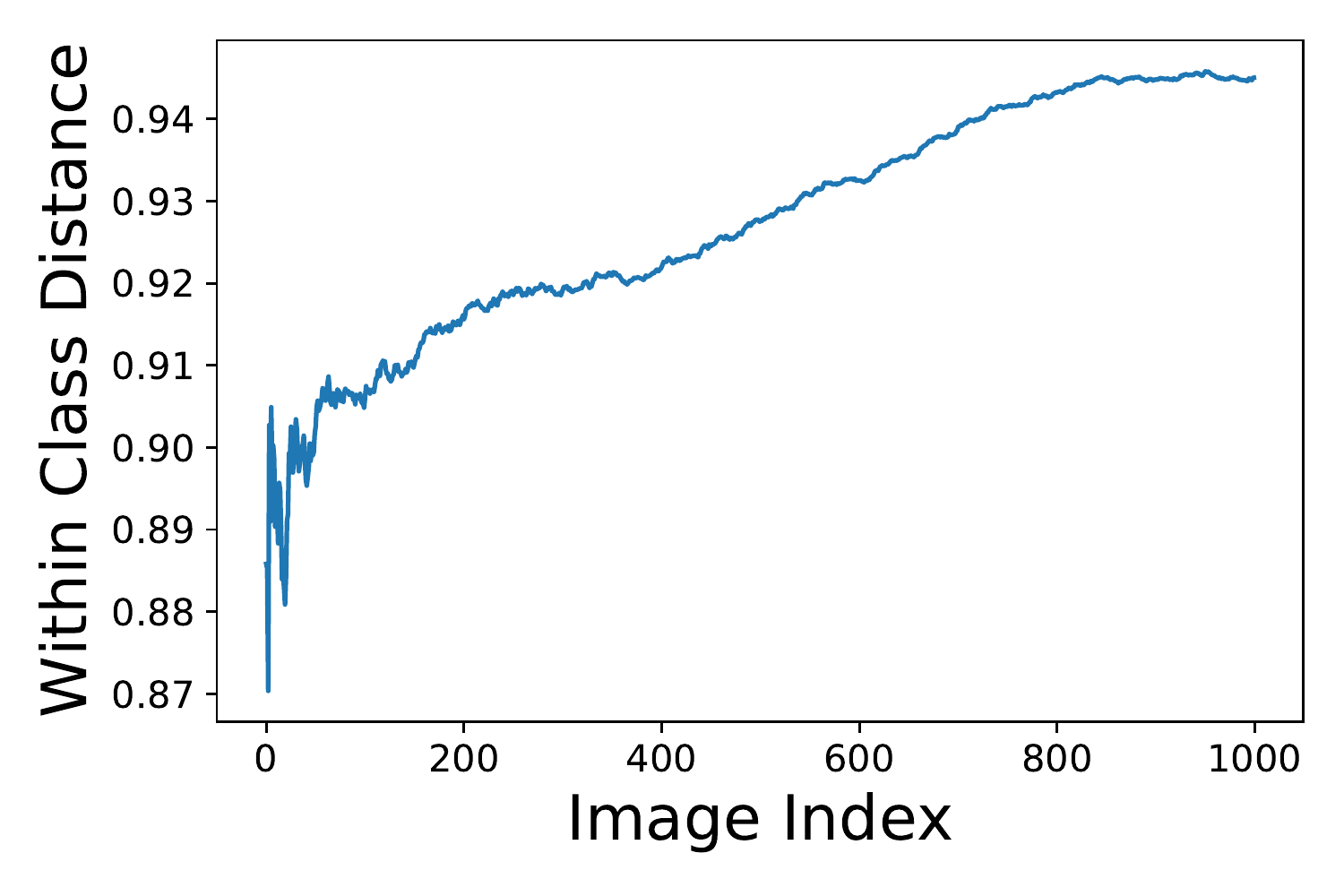}}
    \subfigure[Measure Change Matrix]{\includegraphics[width=0.25\linewidth]{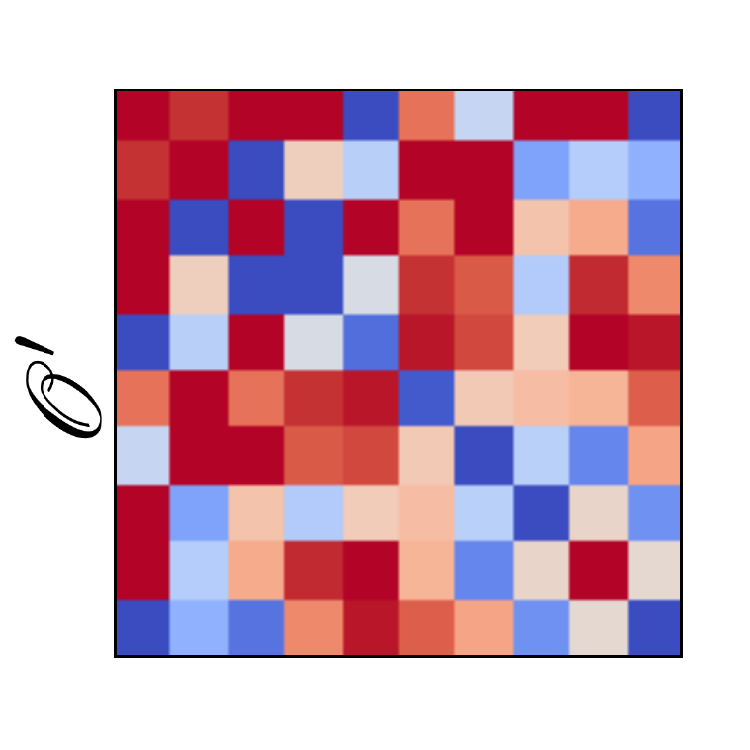}}
    \subfigure[Optimized Measures]{\includegraphics[width=0.32\linewidth]{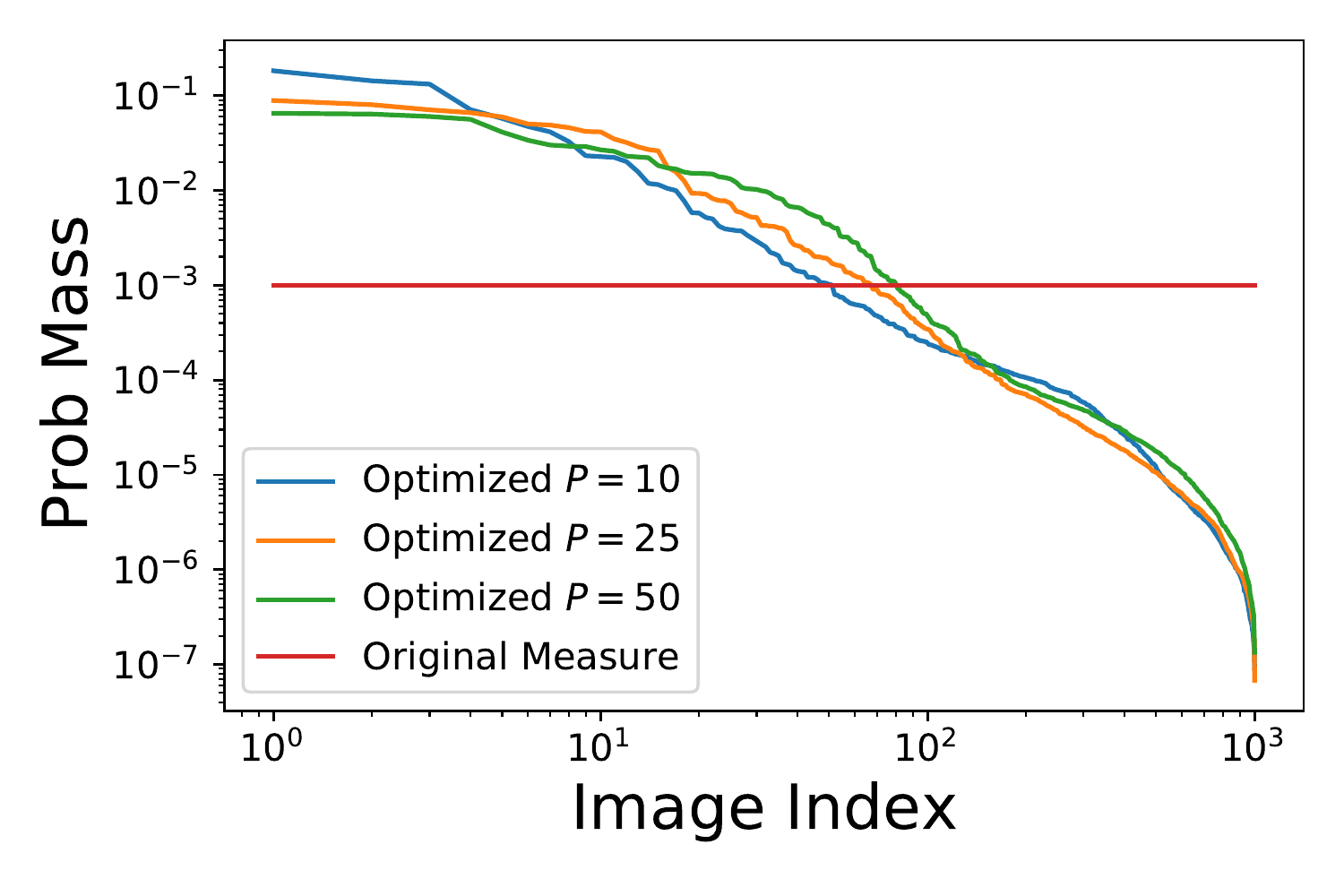}}
    \subfigure[Kernel Regression]{\includegraphics[width=0.32\linewidth]{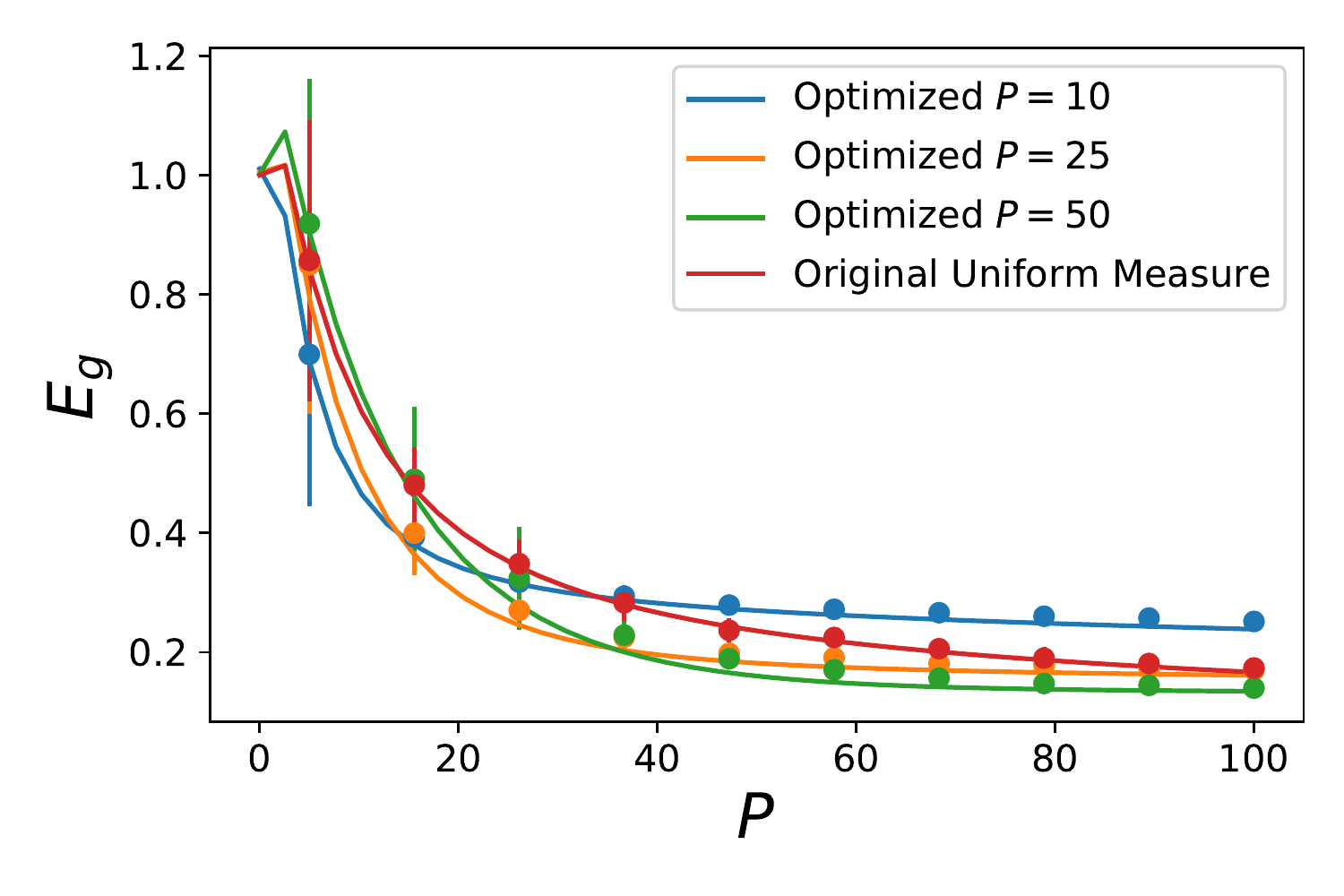}}
    \subfigure[Width 2000 Neural Network]{\includegraphics[width=0.32\linewidth]{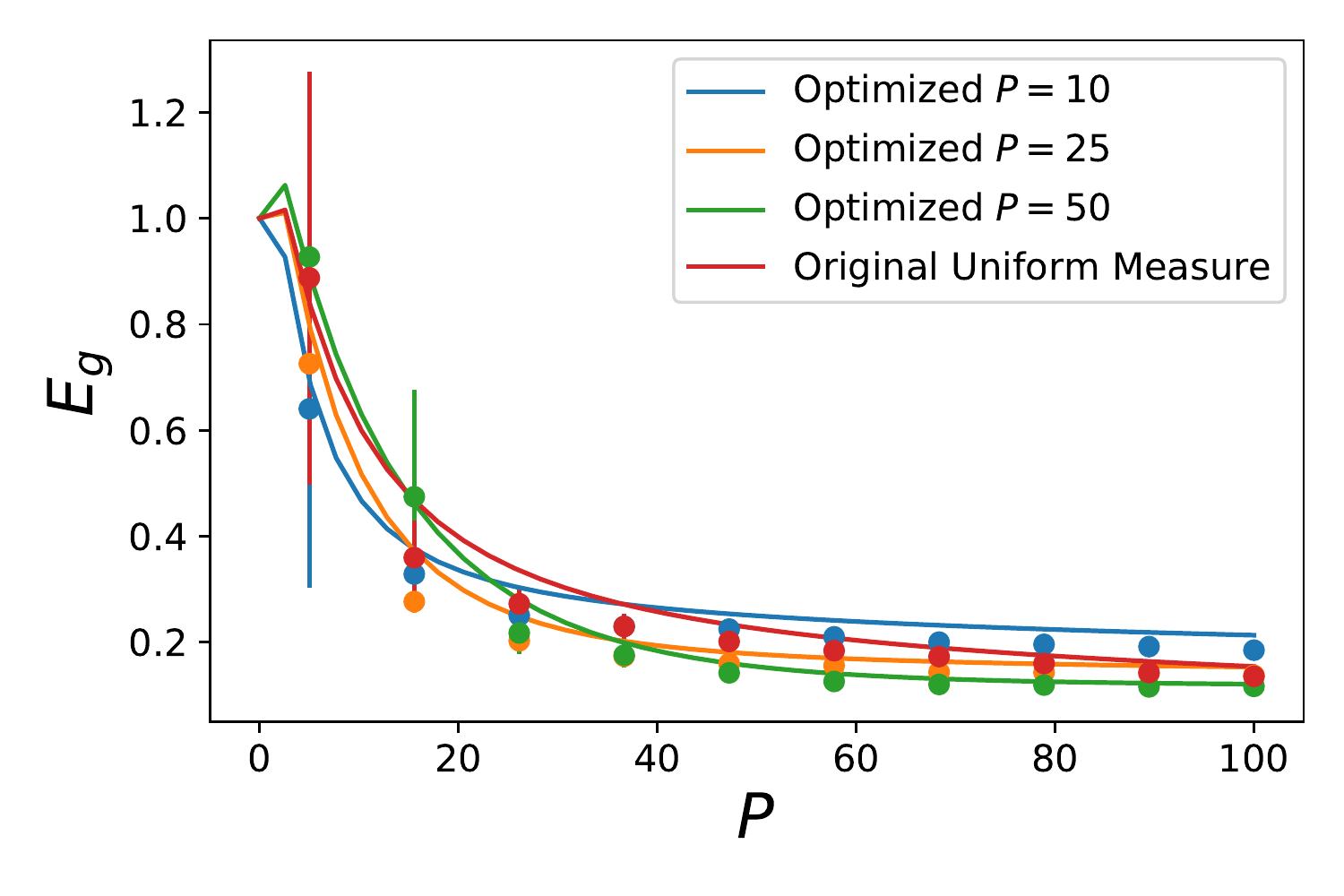}}
    \caption{We study regression with a uniform test distribution on MNIST 8's and 9's. For this fixed task, we optimize our theoretical generalization error expression over training distributions at $P = 30$. This gives a probability distribution over MNIST digits shown in (a). The $30$ images with highest probability mass (b) appear qualitatively more representative of handwritten 8's and 9's than those given the lowest probability mass (c). We plot the theoretical (solid) and experimental (dots) learning curves for the optimized and original uniform training measure, showing that changing the sampling strategy can improve generalization. Error bars display standard deviation over $30$ repeats. (e) After ordering each point by their probability mass on the optimized measure, we calculate the average feature space (in the sense of the kernel) distance to all other points from the same class. This measure rises with image index, indicating that images with higher probability are approximately centroids for each class. (f) The optimized measure induces a non-zero measure change matrix $\mathscr{O}'$. (g) The optimized probability distributions have different shapes for different training budget sizes $P$, with flatter distributions at larger $P$. (h) Kernel regression experiments agree with theory for each of these measures. A measure which performs best at low sample sizes may give sub-optimal generalization at larger sample sizes. (i) The theory also approximates learning curves for finite width $2000$ fully connected neural networks with depth 3, initialized with NTK initialization \cite{jacot2018neural}.}
    \label{fig:mnist_eg_optimization}
\end{figure}

As an example, we fix the test distribution to be uniform MNIST digits for 8's and 9's and optimize over the probability mass of each unique digit to minimize $E_g$, again using a depth $3$ ReLU neural tangent kernel and binary targets $\{-1,1\}$. Our results indicate that it is possible to reduce generalization when the training distribution is chosen differently than the uniform test distribution {(See Section \ref{SI:further_real_data} for extended discussion)}. In \figref{fig:mnist_eg_optimization}, we optimize $E_g$ for $P = 30$ training samples to extract the optimal training distribution on NTK. We observe that the high mass training digits are closer to the center of their class, \figref{fig:mnist_eg_optimization}(e). However we also find that optimizing the distribution for different values of $P$ may cause worse generalization beyond the $P$ used to optimize the distribution \figref{fig:mnist_eg_optimization}(g, h).

We also test our results on neural networks, exploiting a correspondence between kernel regression with the Neural Tangent Kernel, and training infinitely wide neural networks \cite{jacot2018neural}. \figref{fig:mnist_eg_optimization}(i) shows that our theory matches experiments with ReLU networks of modest width $2000$.

\section{Linear Regression: An Analytically Solvable Model}\label{section:main_linear_regression}



\begin{figure}[t]
    \centering
    \subfigure[Varying Training Measure Dimension]{\includegraphics[width=0.37\linewidth]{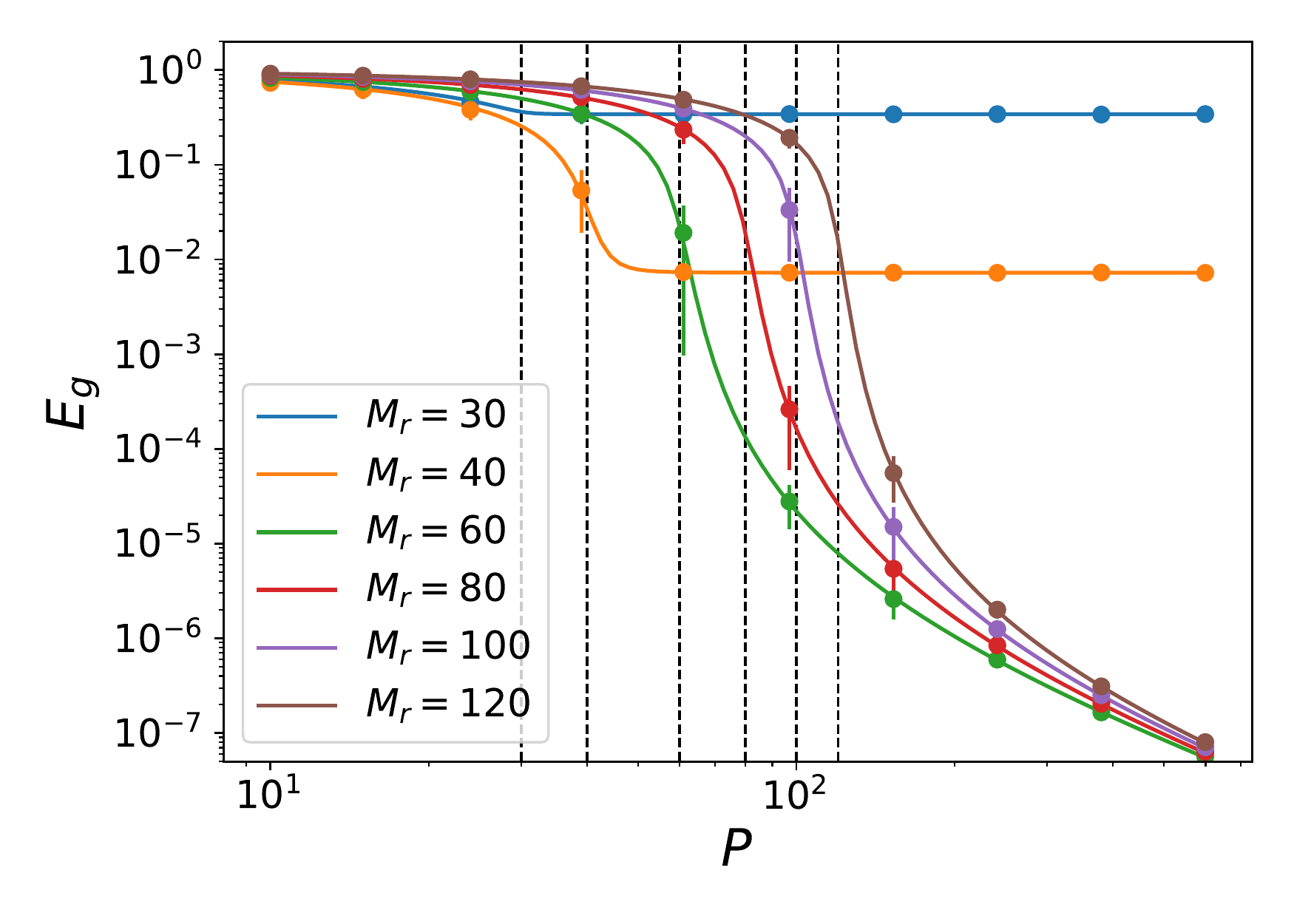}}
    \subfigure[Effect of Training Measure]{\includegraphics[width=0.56\linewidth]{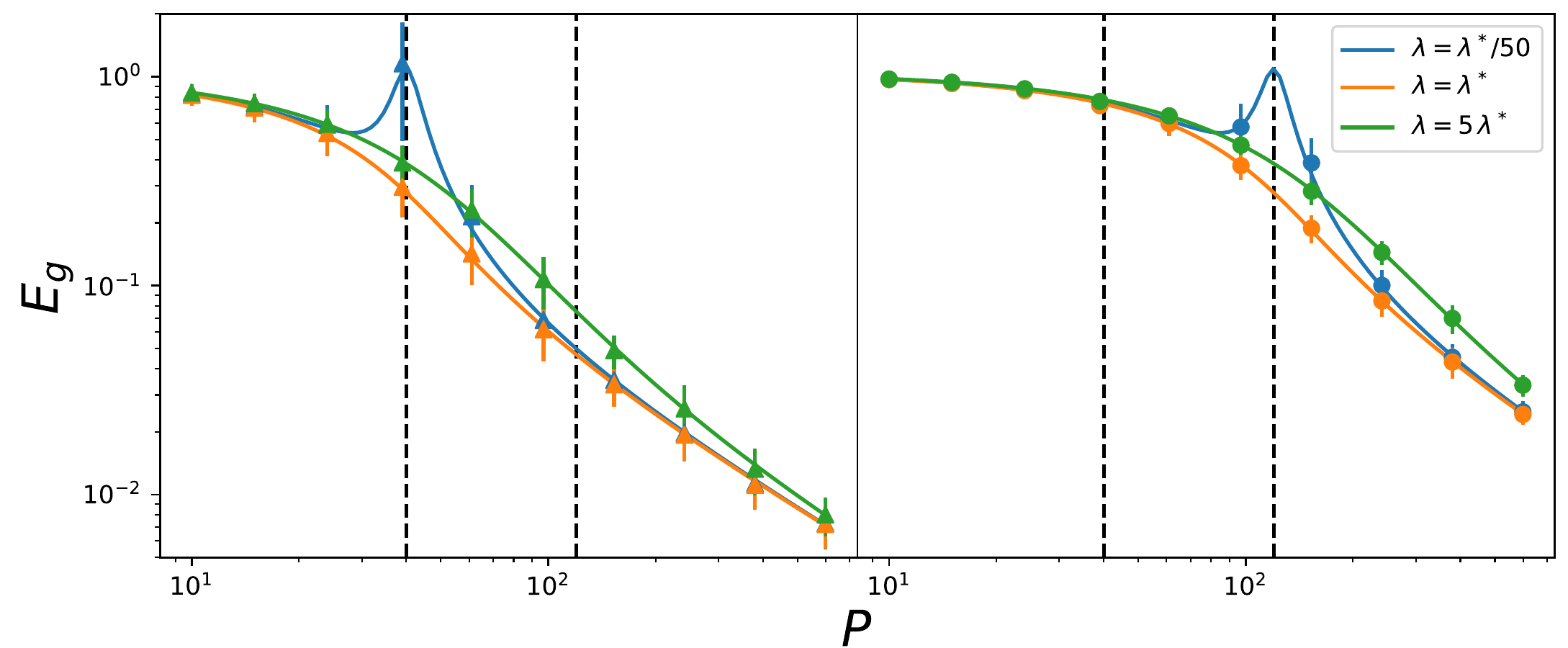}}
    \caption{ (a) Learning curves for linear regression on a target $\bar f = \beta^\top\x$ where $\beta_{\rho \leq  40}\sim\mathcal{N}(0,1)$, $\beta_{60>\rho > 40}\sim\mathcal{N}(0,0.01)$ and $\beta_{\rho>60} = 0$, hence $N = 60$. Input dimension, label noise and ridge parameter are $D = 120$, $\varepsilon^2 = 0$ and $\lambda = 10^{-3}$. Altering the training measure dimensionality below or above $M_{r} = 60$ hurts generalization. For all curves $E_g\to 0$ as $P\to\infty$ except the blue and orange lines at $M_{r} = 30, 40$ where there is irreducible error since some portion of the target coefficients are not learned.
    (b) Same experiment with $\varepsilon^2 = 0.1$ label noise and with varying ridge parameters $\lambda$. Target coefficients are $\beta_{\rho \leq  40}\sim\mathcal{N}(0,1)$ and $\beta_{\rho>40} = 0$ ($N=40$). The variances of training and test distributions are $\sigma^2=\tilde\sigma^2 = 1$. Left panel and right panels are the learning curves for $M_r = 40, 120$, respectively. Dashed vertical lines indicate the number $P = M_r = 40, 120$. Dots are experiment, lines are theory. Error bars represent standard deviation over $30$ trials.
    }
    \label{fig:training_dist_lin_regression}
\end{figure}

Next we study linear regression to demonstrate various interesting OOD generalization phenomena. Consider a linear target function $\bar f(\x) = \beta^\top\x$ where $\x,\beta \in \mathbb{R}^D$ and a linear kernel $K(\x,\x') = \frac{1}{D} \x^\top \x'$ (in this model $M=D$). Data are sampled from zero-mean Gaussians with arbitrary covariance matrices $\C$ and $\tilde\C$ for training and test distributions, respectively.

In this case, the kernel eigenvalue equation can be solved exactly. 
Denoting the eigenvalues and eigenvectors of the covariance matrix of the training distribution as $\C \U = \U \bSigma$ where $\bSigma_{\rho\gamma} = \sigma_\rho^2\delta_{\rho\gamma}$ is the diagonal eigenvalue matrix and $\U$ is an orthogonal matrix with columns being eigenvectors of $\C$, the integral eigenvalue problem becomes $
    \eta_\rho\phi_\rho(\x) = \braket{K(\x,.),\phi_\rho(.)}_{p(\x)}  =  \frac{\sigma_\rho}{D}\u_\rho^\top\x$. Therefore, the normalized eigenfunctions are $\phi_\rho(\x) = \u_\rho^\top\x/\sigma_\rho$ and the eigenvalues are $\eta_\rho = \sigma_\rho^2/D$. 
    The overlap matrix is \begin{align}
    \mathscr{O}_{\rho\gamma } = \braket{\phi_\rho(.),\phi_\gamma(.)}_{\tilde p(\x)} = \bSigma^{-1/2}\U^\top\tilde\C \U \bSigma^{-1/2}.\end{align}
Finally computing the target weights as $\a = \bSigma^{1/2}\U^\top\beta$, we obtain the generalization error \eqref{eq:main_gen_error}:
\begin{gather}\label{eq:gen_err_gaussian}
    E_g = E_g^{0,p(\x)} + \frac{\gamma' - \gamma}{1-\gamma}\varepsilon^2 +  (\kappa D)^2\beta^\top (P\C+\kappa D\I)^{-1} \bigg(\tilde\C-\frac{1-\gamma'}{1-\gamma}\C\bigg)(P\C+\kappa D\I)^{-1}\beta,\nonumber\\
    \gamma = P\Tr \C^2 (P\C+\kappa D\I)^{-2},\; \gamma' = P\Tr \tilde\C \C (P\C+\kappa D\I)^{-2},\nonumber\\
    \kappa = \lambda  + \kappa\Tr \C(P\C+\kappa D\I)^{-1}
\end{gather}
and
\begin{align}
    E_g^{0,p(\x)} = \frac{\gamma}{1-\gamma}\varepsilon^2 + \frac{(\kappa D)^2}{1-\gamma}\beta^\top\C(P\C+\kappa D \I)^{-2}\beta.
\end{align}
As a consistency check, we note that the generalization error is minimized ($E_g = 0$) when $\tilde\C = 0$ corresponding to a Dirac measure at the origin. This makes sense since target function at origin is $0$ and the estimator on the test distribution is also $0$. 

Next, we consider a  diagonal covariance matrix $\tilde\C = \tilde \sigma^2\I$ for test distribution and
\begin{align}
    \C = \diag\, (\underbrace{\sigma^2,\dots\sigma^2}_{M_r},\underbrace{0,\dots 0}_{D-M_r})
\end{align}
for training distribution to demonstrate how training distribution affects generalization in a simplified setting. The integer $M_r$ corresponds to the rank of the training measure's covariance which is also the number of non-zero kernel eigenvalues for the linear kernel. Furthermore we take the target function to have power only in the first $N$ features ($\beta_{\rho>N} = 0$) where we adopt to the normalization $\sum_{\rho = 1}^N\beta_\rho^2 = 1$. Thus, the target does not depend on the $D-N$ remaining dimensions $x_{\rho>N}$ and we study how compressing some directions in training distribution influences generalization. 

In this case, the self-consistent equation for $\kappa$ becomes exactly solvable. The generalization error reduces to:
\begin{align}\label{eq:eg_diagonal_corvariance}
    E_g &= \tilde\sigma^2\bigg(\frac{\varepsilon^2}{\sigma^2} \frac{\alpha}{(\kappa' + \alpha)^2 - \alpha}+ \frac{\kappa'^2}{(\kappa' + \alpha)^2 - \alpha}\sum_{\rho=1}^{M_r}\beta_\rho^2 + \sum_{\rho=M_r+1}^{N}\beta_\rho^2 \bigg),
\end{align}
where $\kappa' = \frac{\kappa}{\sigma^2 M_r/D} =  \frac{1}{2}\big[(1+\tilde\lambda-\alpha) + \sqrt{(1+\tilde\lambda+\alpha\big)^2-4\alpha}]$, $\tilde\lambda = \frac{\lambda}{\sigma^2 M_r/D}$ and $\alpha = P/M_r$. This result matches that of \cite{canatar2020spectral} analyzing in-distribution generalization error when $M_r=N=D$ and $\sigma^2 = \tilde \sigma^2$. We identify $\tilde\lambda$ as an {\it effective regularization} parameter as it assumes the role $\lambda$ plays for in-distribution generalization (compare to Eq. 7 of \cite{canatar2020spectral}).

First, we note that the generalization linearly scales with the variance of the test distribution. This is due to the linearity of the target and the estimator; any mismatch between them is amplified when the test points are further away from the origin.

Next, when the dimension of training measure is smaller than the dimension of the target function, $M_r < N$, there is an irreducible error due to the fact that kernel machine cannot fit the $N-M_r$ dimensions which are not expressed in the training data (\figref{fig:training_dist_lin_regression}a). However in certain cases choosing $M_r < N$ may help generalization if the data budget is limited to $P \sim M_r$ since the learning happens faster compared to larger $M_r$. As illustrated in \figref{fig:training_dist_lin_regression}a, if the target power is distributed such that $\sum_{\rho=N'+1}^{N}\beta_\rho^2 \ll \sum_{\rho=1}^{N'}\beta_\rho^2$, choosing $M_r = N'$ can be a better strategy despite the irreducible error due to unexplored target power. In \figref{fig:training_dist_lin_regression}a, we picked $N = 60$ and $N'=40$ which is the number of directions target places most of its power on. If the data budget was limited to $P<N$, choosing $M_r = N'$ (orange curve in \figref{fig:training_dist_lin_regression}a) performs better than $M_r = N$ (green curve) although for $P \geq N$ the $M_r < N$ curve has irreduible error while the $M_r=N$ does not. Reducing $M_r$ below $N'$ (blue curve in \figref{fig:training_dist_lin_regression}a) on the other hand does not provide much benefit at small $P$.

Next, we observe that $\kappa'$ is a monotonically decreasing function of $\alpha$ and shows a sharp decrease near $\alpha \approx 1 + \tilde\lambda$ which also implies a sharp change in generalization error. In fact when $\tilde\lambda=0$, $\kappa'$ becomes zero at $\alpha = 1$ and its first derivative diverges implying either a divergence in generalization error due to label noise 
or vanishing of $E_g$ in the absence of noise (Section \ref{SI:linear_regression}). {When the labels are noisy, this non-monotonic behavior $\alpha = 1$ in $E_g$ (called sample-wise double-descent) signals the over-fitting of the labels beyond which the kernel machine is able to average over noise and converge to the true target function \cite{belkin2019reconciling,nakkiran2019deep,krogh,Hertz_1989, montanari2019surprises,nakkiran2019moredata,nakkiran2020optimal,dascoli2020triple,liang_isometry,chen2020multipledesign,canatar2020spectral}.} 
Hence reducing $M_r$ means that this transition occurs earlier in the learning curve ($P \sim (1+\tilde\lambda)M_r$) which implies that less training samples are necessary to estimate the target function well. Intuitively, sampling from an effectively higher dimensional space increases the necessary amount of training data to fit the modes in all directions. In \figref{fig:training_dist_lin_regression}(b), we demonstrate this prediction for a linear target and obtain perfect agreement with experiment.

Finally, we show that increasing effective regularization $\tilde\lambda$, which is controlled by the training distribution ($\sigma^2$, $M_r$), leads to the suppression of double-descent peak, hence avoiding possible non-monotonicities. However large $\tilde\lambda$ leads to slower learning creating a trade-off. Results from previous studies \cite{nakkiran2020optimal,canatar2020spectral} have shown the existence of an optimal ridge parameter in linear regression minimizing the generalization error for all $P$. {Minimizing \eqref{eq:eg_diagonal_corvariance} with respect to $\lambda$, we find that the optimal ridge parameter is given by:}
\begin{align}
    \lambda^* = \frac{M_r}{D}\varepsilon^2.
\end{align}
In \figref{fig:training_dist_lin_regression}(b), we show that choosing optimal ridge parameters indeed mitigates the double descent and results in the best generalization performance.

\section{Further Results}

In SI, we present other analytically solvable models and further analysis on distribution mismatch for real data applications.

\begin{itemize}[itemsep = 0pt, leftmargin = 0pt]
    \item In Section \ref{SI:further_real_data}, we analyze the gradient descent/ascent procedures performed in Section \ref{sec:real_datasets} and provide further experiments to motivate how our theory can be used to find mismatched train/test distributions which improve generalization. We also apply our theory to adversarial attacks during testing.
    
    \item In Section \ref{SI:linear_regression}, we study a more general linear regression model with diagonal overlap matrix where train/test distributions, target function and number of directions kernel expresses vary. When the target has out-of-RKHS components, we show how distribution shifts may help in generalization.
    \item In Section \ref{SI:ntk}, we apply our theory to rotation invariant kernels such as the Gaussian RBF kernel and NTK \cite{jacot2018neural} acting on spherical data in high-dimensions. We examine how a  mismatched sphere radius affects the generalization error in the limit $P, D\to\infty$ similar to the case solved in \cite{bordelon2020spectrum,canatar2020spectral}. 
    
    \item In Section \ref{SI:interpolation_extrapolation}, we study interpolation versus extrapolation in linear regression and regression in Fourier space by studying  rectangular distributions with different ranges for one-dimensional inputs.
\end{itemize}

\section{Discussion}

Interest in kernel methods has recently surged due to their profound connections to deep neural networks \cite{jacot2018neural,belkin2018understand,lee2017deep,lee2019wide}, and the replica method of statistical physics proved useful in studying its generalization properties, inductive biases and non-monotonic learning curves \cite{bordelon2020spectrum,canatar2020spectral,sompolinsky1999statistical,loureiro2021capturing,sompolinsky1992examples,optimalperceptron,dascoli2020triple, dascoli2020double}. Along the lines of these works, we analyzed generalization performance of kernel regression under distribution mismatch and obtained analytical expressions which fit experiments perfectly including those with wide neural networks.  We demonstrated that our formula can be used to optimize over training distribution on MNIST which revealed that certain digits are better to sample more often than others. We considered several analytically solvable models, particularly linear regression for which we demonstrated how dimensional reduction of the training distribution can be helpful in generalization, including shifting of a double-descent peak. 
Our theory brings many insights about how kernel regression generalizes to unseen distributions. 


In this work, we focused on generalization error for specific mismatched training and test distributions, as in \cite{gonzalez2015mismatched,kouw2019review}. One could, instead, consider a set of possible test distributions that the model could encounter, and assess out-of-distribution generalization error based on the worst performance on that set. This distributional robustness {approach} \cite{wald1945statistical} has been the focus of most previous studies \cite{tsipras2018robustness,sagawa2019distributionally,ilyas2019adversarial,sagawa2020investigation} {where one can train a model by minimizing a more general empirical risk on a subset of all test distributions using robust optimization techniques \cite{ben2009robust}.} 
More recent approaches aim to learn invariant relationships across all distributions \cite{arjovsky2019invariant, arjovsky2020out,krueger2020outofdistribution}. It will be interesting to see if our approach can be generalized to ensembles of training and test distributions.  
 
While our theory accurately describes experiments, it has limitations. First, using it to optimize for a training distribution requires the knowledge of exact test distribution which could not be available at the time of training. Second, the theory requires an eigendecomposition of the kernel which is computationally costly for large datasets. {This problem, however, can potentially be solved by using stochastic methods to compute the kernel regression solution \cite{dai2014scalable}.} Third, our theory uses the replica theory \cite{mezard1987spin}, which is not fully rigorous. Fourth, if to be used to describe neural networks, our theory's applicability is limited to their kernel regime.  


\bibliographystyle{unsrt} 
\bibliography{bibliography}


\newpage 

\appendix

\setcounter{section}{0}
\setcounter{equation}{0}
\setcounter{figure}{0}
\setcounter{page}{1}
\renewcommand{\theequation}{SI.\arabic{equation}}
\renewcommand{\thefigure}{SI.\arabic{figure}}
\renewcommand{\thesection}{SI.\arabic{section}}

\numberwithin{equation}{section}
\numberwithin{figure}{section}

\begin{center}
    \Large\textbf{Supplemental Information for \\ ``Out-of-Distribution Generalization in Kernel Regression''}
\end{center}

\section{Calculation of Generalization Error}\label{SI:calculation}

\subsection{Problem Formulation} 

We consider a probability distribution $p(\x)$ on the input space $\mathcal{X} \subset \mathbb{R}^D$ and an orthonormal basis $\{\phi_\rho(\x)\}|_{\rho = 0}^M$ ($M$ is typically infinite) spanning the space of square integrable functions $L^2(\mathcal{X})$ such that any square integrable function can be expanded as:
\begin{align}
    f(\x) = \sum_{\rho = 1}^M a_\rho \phi_\rho(\x),\quad \braket{f,f}_{p(\x)} = \sum_{\rho = 1}^M a_\rho^2 < \infty,
\end{align}
where $\braket{f,f}_{p(\x)}$ denotes the $L^2$ norm of the function. 

A reproducing kernel Hilbert space (RKHS) $\mathcal{H}$ is a Hilbert space endowed with an inner product $\braket{.,.}_\mathcal{H}$ where evaluation operator is continuous, mapping any function $f\in\mathcal{H}$ to its value at $f(\x)$ \cite{RasmussenWilliams,scholkopf2002learning}:
\begin{align}
    f(\x) = \braket{f(.), K(.,\x)}_\mathcal{H}\quad \forall f \in \mathcal{H}.
\end{align}
Here the so-called reproducing kernel $K: \mathcal{X}\times\mathcal{X}\to\mathbb{R}$ is a positive-definite function whose partial evaluation $K(.,\x)$ itself belongs to $\mathcal{H}$. It can be characterized by the integral operator $T_K$:
\begin{align}
    [T_K f](x) = \int d\x' p(\x') K(\x, \x') f(\x'),
\end{align}
where $T_K$ has spectral decomposition $[T_K\phi_\rho](\x) = \eta_\rho\phi_\rho(\x)$ for $\rho=1,\dots,M$. Then \textit{Mercer's theorem} allows the following kernel representation in terms of orthonormal functions $\{\phi_\rho(\x)\}|_{\rho = 0}^M$:
\begin{align}\label{mercer}
	K(\x,\x') &= \sum_{\rho = 1}^N \eta_\rho\phi_{\rho}(\x)\phi_{\rho}(\x') = \bPhi(\x)^\top\bLambda\bPhi(\x') = \bPsi(\x)^\top \bPsi(\x),\\\nonumber
	\psi_{\rho}(\x)&\equiv\sqrt{\eta_\rho}\phi_{\rho}(\x) ,\quad \braket{\psi_{\rho}(\x),\psi_{\gamma}(\x)}_\mathcal{H} = \delta_{\rho\gamma},
\end{align}
where we defined the diagonal eigenvalue matrix $\bLambda_{\rho\gamma} = \eta_\rho\delta_{\rho\gamma}$, $M$-dimensional vector $\big(\bPhi(\x)\big)_\rho = \phi_\rho(\x)$ and features $\psi_\rho$. With this representation Hilbert inner product $\braket{.,.}_\mathcal{H}$ reduces to:
\begin{align}\label{eq:hilbert_inner_product}
    \braket{f,g}_\mathcal{H} \equiv \sum_{\rho=1}^M \frac{a_\rho b_\rho}{\eta_\rho},
\end{align}
for two functions $f(\x) = \a^\top\bPhi(\x)$ and $g(\x) = \b^\top\bPhi(\x)$. A function belongs to this RKHS only if its Hilbert norm is finite:
\begin{align}
    \norm{f}_\mathcal{H}^2 \equiv \braket{f,f}_\mathcal{H} = \sum_{\rho = 1}^M \frac{a_\rho^2}{\eta_\rho} < \infty.
\end{align}
Note that the kernel does not have to represent all $M$-features meaning that its eigenvalues may truncate at some integer $N < M$: $\eta_{\rho > N} = 0$. If this is the case, then functions which have power on modes $\rho>N$ are out-of-RKHS functions since they have infinite Hilbert norm.

\vskip 0.5em

Given a finite training data $\mathcal{D}=\{\x^\mu, y^\mu\}_{\mu = 1}^P$ where inputs are drawn from probability distribution $p(\x)$, we wish to study kernel regression on the RKHS $\mathcal{H}$. First, we assume that the labels are generated by a target function $\bar f(\x)$ with additive noise:
\begin{align}
  y^\mu = \bar f(\x^\mu) + \epsilon^\mu,
\end{align}
where $\mathbb{E}[\epsilon^\mu\epsilon^\nu] = \varepsilon^2\delta_{\mu\nu}$ are independent for each training sample. In general, the target function does not have to be in the RKHS and the out-of-RKHS components need to be treated separately when $\eta_\rho = 0$ for some $\rho$. However we find that taking the limit $\eta_\rho\to 0$ at the end of the calculation yields the correct expressions for out-of-RKHS cases hence we keep all $\eta_\rho \neq 0$ for now. Expanding the target function in terms of the eigenfunctions with respect to $p(\x)$:
\begin{align}
    y^\mu = \bar\a^\top\bPhi(\x) + \epsilon^\mu = \bar\w^\top\bPsi(\x^\mu) + \epsilon^\mu.
\end{align}
The problem of interest is the minimization of the energy function $H[f;\mathcal{D}]$ with respect to functions $f\in\mathcal{H}$:
 \begin{align}
	\w^* = \text{argmin}_{\w \in \R^{M}}H(\w;\mathcal{D}),\quad H(\w;\mathcal{D}) \equiv &\frac{1}{2\lambda} \sum_{\mu=1}^P\left(\bPsi(\x^\mu)\cdot(\bar\w-\w) +\epsilon^\mu\right)^2+ \frac{1}{2}\Vert\w\Vert_2^2,
\end{align}
Then the resulting estimator becomes:
\begin{align}
    f^*(\x;\mathcal{D}) = {\w^*}^\top\bPsi(\x),
\end{align}
Note that the estimator is always in the RKHS, meaning it should not depend on eigenfunctions $\phi_\rho(\x)$ if the corresponding eigenvalue $\eta_\rho = 0$.

\vskip 0.5em

The estimator above depends on the particular choice of a training set and it is difficult to obtain an analytical expression for it. Hence we would like to compute the average case estimator which only depends on the input distribution, size of the dataset, target function and the hypothesis class $\mathcal{H}$ but not the individual training samples. Next we discuss how to perform dataset averaging for kernel regression using methods from statistical physics.

\subsection{Replica Calculation for Generalization}

We would like to calculate the observables $\mathcal{O}[f^*]$ of the estimator $f^*(\x)$ averaged over the training dataset $\mathcal{D}$. These observables include the generalization and training errors. For our purposes, we only need to calculate the mean and variance of the estimator which completely determines the generalization error. To perform this calculation, we introduce the following partition function:
 \begin{equation}
	Z[\bxi,\bchi] = \int d\w e^{-\beta H(\w;\mathcal{D})+\beta \bxi \cdot \w + \frac{\beta}{2}\bchi^\top \w \w^\top\bchi},
\end{equation}
where $\beta$ is inverse temperature and $\bxi,\bchi$ are source terms to compute expectation values of the estimator weights. The partition function represents a probability distribution over all possible estimators $\w$ and as $\beta\to\infty$ it concentrates around the kernel regression solution $\w^*$.

\vskip 0.5em

We can calculate the dataset averaged estimator $f^*(\x)$ and its correlation function via:
 
\begin{align}
    \sqrt{\eta_\alpha} \mathbb{E}_\mathcal{D}{w^*_\alpha(\mathcal{D})} &= \lim_{\beta\to\infty}\left.\frac{1}{\beta}\frac{\partial}{\partial\xi'_\alpha}\mathbb{E}_\mathcal{D}{\log Z[\bxi,\bchi]}\right|_{{\bxi,\bchi} = 0}\nonumber\\
    \sqrt{\eta_\alpha\eta_\beta} \mathbb{E}_\mathcal{D}[w^*_\alpha(\mathcal{D}) w^*_\beta(\mathcal{D})] &= \lim_{\beta\to\infty}\left.\frac{1}{\beta}\frac{\partial^2}{\partial\chi'_\alpha\partial\chi'_\beta}\mathbb{E}_\mathcal{D}{\log Z[\bxi,\bchi]}\right|_{{\bxi,\bchi} = 0},
\end{align}

where the primed quantities are defined as $\bxi = \bLambda^{1/2}\bxi'$ and $\bchi = \bLambda^{1/2}\bchi'$. Hence one can read out:
 
\begin{align}
    \mathbb{E}_\mathcal{D}{f^*(\x,\mathcal{D})} &= \sum_\alpha  \mathbb{E}_\mathcal{D}{w^*_\alpha(\mathcal{D})} \sqrt{\eta_\alpha}\phi_\alpha(\x)\nonumber\\
    \mathbb{E}_\mathcal{D}[{f^*(\x,\mathcal{D})f^*(\x',\mathcal{D})}] &= \sum_{\alpha\beta} \mathbb{E}_\mathcal{D}[w^*_\alpha(\mathcal{D}) w^*_\beta(\mathcal{D})]\sqrt{\eta_\alpha\eta_\beta} \phi_\alpha(\x)\phi_\beta(\x').
\end{align}

However, computing the average of $\log Z$ over all possible training samples and noises is a challenging task {due to the integrals of logarithms.} This is where we resort to replica trick which replaces averaging $\log Z$ with averaging $Z^n$, n-times \textit{replicated} partition function:
 \begin{align}
	&\mathbb{E}_\mathcal{D}{\log Z } = \lim_{n\rightarrow 0}\frac{\mathbb{E}_\mathcal{D}{Z^n}-1}{n}.
\end{align}
{The calculation of $\mathbb{E}_\mathcal{D}{Z^n}$ is done for integer $n$ and then analytically continued to real numbers to perform the limit. Despite being non-rigorous, the replica method has proven powerful and predictive in the study of disordered systems as well as neural networks and machine learning (see \cite{mezard1987spin,spinGlassReview,advani2013statistical} for reviews).} 

Plugging in the eigenfunction expansions derived above, $Z^n$ becomes:
 \begin{alignat}{2}
	\mathbb{E}_\mathcal{D}{Z^n} &= 	e^{-\frac{n\beta}{2}(\bar\W-\bxi)^\top\bar\w }\int \bigg(\prod_{a=1}^n d\w^a\bigg) && e^{-\frac{\beta}{2}\sum_{a=1}^n{\w^a}^\top(\I - \bchi\bchi^\top)\w^a -\beta\bar\W^\top\sum_{a=1}^n\w^a}\nonumber\\
	& &&\times\Braket{ e^{-\frac{\beta}{2\lambda} \sum_{a=1}^n\big(\w^a\cdot\bPsi(\x) +  \epsilon^a\big)^2}}_{\x\sim p(\x),\{\epsilon^a\}}^P,
\end{alignat}
where we shifted $\w^a \to \w^a + \bar\w$ and defined $\bar \W = (\bar\w-\bxi)-(\bchi^\top\bar\w)\bchi$ for notational convenience. A crucial step is to compute the expectation value over possible realizations of training sets which behave as quenched disorders in our system. Our strategy to perform this quenched average is approximating the exponent as a Gaussian random variable defined by $q^a=(\w^a-\bar \w)\cdot\bPsi(\x) +\epsilon^a$ whose second-order statistics is given by:
 \begin{align}\label{order_params}
	\bmu^a \equiv \braket{q^a} &= 0, \nonumber \\
	\C^{ab}\equiv\braket{q^a q^b} &= (\w^a-\bar \w)^\top \braket{\bPsi(\x)\bPsi(\x)^T}(\w^b-\bar \w)+\braket{\epsilon^a\epsilon^b}=(\w^a-\bar \w)^\top \bLambda(\w^b-\bar \w) + \bSigma^{ab},
\end{align}
where $\bSigma = (\varepsilon^2 +\norm{\bar\a}_2^2) \1\1^\top$ is the covariance matrix of noise across replicas. We assumed that the kernel does not include a constant mode so that $\bmu^a = 0$. Noticing that $q^a$ is a summation of many uncorrelated random variables ($\left<   \psi_\rho(\mathbf{x}),\psi_{\rho'}(\mathbf{x})\right>_{p(\x)}= \eta_\rho \delta_{\rho\rho'}$) and a Gaussian noise, we approximate the probability distribution of $q^a$ by a multivariate Gaussian with its mean and covariance given by \eqref{order_params}:
 \begin{equation}
P(\{q^a\}) = \frac{1}{\sqrt{(2\pi)^n\det(\C)}}\exp\bigg(-\frac{1}{2}\sum_{a,b}q^a\big(\C^{ab}\big)^{-1}q^b\bigg),
\end{equation}
The Gaussian approximation proves accurate given the excellent match of our theory to simulations. This reduces the average over quenched disorder to:
 \begin{equation}
	\begin{split}
	\Braket{ e^{-\frac{\beta}{2\lambda} \sum_{a=1}^n\big((\w^a-\bar \w)\cdot\bPsi(\x)+\epsilon^a\big)^2}}_{\x,\{\epsilon^a\}}&\approx \int\,\{dq^a\}P(\{q^a\})\exp\bigg(-\frac{\beta}{2\lambda} \sum_{a=1}^n(q^a)^2\bigg)\\
	& = \exp\bigg(-\frac{1}{2}\log\det(\I+\frac{\beta}{\lambda}\C)\bigg).
	\end{split}
	\end{equation}
Combining everything together, the averaged replicated partition function becomes:
 \begin{equation}
	\mathbb{E}_\mathcal{D}{Z^n} = e^{-\frac{n\beta}{2}(\bar\W-\bxi)^\top\bar\w }\int \bigg(\prod_{a=1}^n d\w^a\bigg) e^{-\frac{\beta}{2}\sum_{a=1}^n{\w^a}^\top(\I- \bchi\bchi^\top)\w^a -\beta\bar\W^\top\sum_{a=1}^n\w^a-\frac{P}{2}\log\det(\I+\frac{\beta}{\lambda}\C)},
\end{equation}
Using the definitions \eqref{order_params}, we insert the following identity to the integral:
 \begin{align}
	1 = \bigg(\frac{i{P}}{2\pi}\bigg)^{\frac{n(n+1)}{2}}\int &\bigg({\prod_{a\geq b}}d\C^{ab}d\hat \C^{ab}\bigg)\exp\Bigg[-{P}{\sum_{a\geq b}}\hat \C^{ab} \bigg(\C^{ab} - {\w^a}^\top\bLambda \w^b - \bSigma^{ab}\bigg)\Bigg]
\end{align}
	Here, the integral over $\hat\C$ runs over the imaginary axis and we explicitly scaled conjugate variables by ${P}$. Then defining:
 \begin{align}
	G_E &=\frac{1}{2}\log\det(\I +\frac{\beta}{\lambda}\C),\\
	e^{-G_S} &= \int \bigg(\prod_{a=1}^n d\w^a\bigg)\exp(-\frac{\beta}{2}{\sum_{a\geq b}}{\w^a}^\top\bigg(\big(\I - \bchi\bchi^\top\big)\I^{ab}-\frac{2{P}}{\beta}\bLambda\hat\C^{ab}\bigg)\w^b -\beta\bar\W^\top\sum_{a=1}^n\w^a),
\end{align}
we obtain:
 \begin{align}\label{eq:rawAction}
	\mathbb{E}_\mathcal{D}{Z^n} = e^{\frac{n(n+1)}{2}\log(\frac{i{P}}{2\pi})-\frac{n\beta}{2}(\bar\W-\bxi)^\top\bar\w }\int &\bigg({\prod_{a\geq b}} d\C^{ab}d\hat \C^{ab}\bigg)\exp\left[-{P}{\sum_{a\geq b}}\hat \C^{ab}(\C^{ab}-\bSigma^{ab})-PG_E- G_S\right].
\end{align}
Next, we need to evaluate the integral in $G_S$. First we would like to express the ordered sum ${\sum_{a\geq b}}{\w^a}^\top\big(\big(\I-\bchi\bchi^\top\big)\I^{ab}-\frac{2{P}}{\beta}\bLambda\hat\C^{ab}\big)\w^b$ as an unordered sum over $a,b$. Note that
 
\begin{align}
    &{\sum_{a, b}}{\w^a}^\top\bigg(\big(\I-\bchi\bchi^\top\big)\I^{ab}-\frac{2{P}}{\beta}\bLambda\hat\C^{ab}\bigg)\w^b\nonumber\\
    &=2{\sum_{a\geq b}}{\w^a}^\top\bigg(\big(\I-\bchi\bchi^\top\big)\I^{ab}-\frac{2{P}}{\beta}\bLambda\hat\C^{ab}\bigg)\w^b - {\sum_{a, b}}{\w^a}^\top\bigg(\big(\I-\bchi\bchi^\top\big)\I^{ab}-\frac{2{P}}{\beta}\bLambda\diag(\hat\C)^{ab}\bigg)\w^b\nonumber
\end{align}

Hence, we obtain:
 
\begin{align}
    {\sum_{a\geq b}}{\w^a}^\top\bigg(\big(\I-\bchi\bchi^\top\big)\I^{ab}-\frac{2{P}}{\beta}\bLambda\hat\C^{ab}\bigg)\w^b = {\sum_{a, b}}{\w^a}^\top \X^{ab}\w^b,\nonumber
\end{align}

where we defined:
 
\begin{align}
    \X^{ab} = \big(\I-\bchi\bchi^\top\big)\I^{ab}-\frac{{P}}{\beta}\bLambda \big(\hat\C+\diag(\hat\C)\big)^{ab}.
\end{align}

In order to evaluate the Gaussian integral, we will cast the function and target weights into an $nM$ dimensional vector:
 
\begin{align}
    \w &= \begin{pmatrix} \w^1 & \w^2 & .. & \w^a& ..  & \w^n \end{pmatrix}_{nM\times 1}\nonumber\\
    \bar\W_{\otimes n} &= \begin{pmatrix} \bar\W& \bar\W& \dots & \bar\W\end{pmatrix}_{nM\times 1}
\end{align}

Furthermore, we introduce the $nM\times nM$ matrix $\X$ as:
 
\begin{align}
    \X = \begin{pmatrix} \X^{11} & \X^{12} & \dots & \dots & \X^{1n} \\
    \X^{21} & \X^{22} &  \hdots & \hdots & \X^{2n}\\
    \vdots & \vdots & \ddots & & \vdots\\
    \vdots & \dots&  \X^{ab} & \ddots  & \vdots\\
    \X^{n1} & \dots & \dots & \dots & \X^{nn}
    \end{pmatrix}_{nM\times nM}
\end{align}

Finally, we denote the integration measure as $\mathcal{D}\w = \prod_{a,\rho}dw_\rho^a$. With these definitions, $G_S$ becomes:
 
\begin{align}
    e^{-G_S} = \int \mathcal{D}\w\, e^{-\frac{\beta}{2} \w^\top \X \w - \beta {\bar\W_{\otimes n}}^\top \w}
\end{align}

Hence, we turned the integral in $G_S$ to a simple Gaussian integral. The result is:
 
\begin{align}
    e^{-G_S} = \bigg(\frac{2\pi}{\beta}\bigg)^{\frac{nM}{2}}\big(\det\X\big)^{-\frac{1}{2}}\exp(\frac{\beta}{2}{\bar\W_{\otimes n}}^\top\X^{-1}{\bar\W_{\otimes n}}).
\end{align}

Now the integral in \eqref{eq:rawAction} can be evaluated using the method of steepest descent. In \eqref{eq:rawAction}, we see that all the terms in the exponent is $\mathcal{O}(n)$. Furthermore, we will use ${P}$ as the saddle point parameter going to infinity with proper scaling. Therefore, defining the following function:
 \begin{align}\label{eq:actions}
	S[\C,\hat\C,\bmu,\hat\bmu] &= \frac{1}{n}{\sum_{a\geq b}}\hat \C^{ab}(\C^{ab}-\bSigma^{ab})+ \frac{1}{n{P}}\bigg(PG_E+G_S + \frac{n\beta}{2}(\bar\W-\bxi)^\top\bar\w\bigg)\nonumber\\
	G_E &=\frac{1}{2}\log\det(\I +\frac{\beta_K}{\lambda}\C)\nonumber\\
	G_S &= \frac{1}{2} \log\det \X-\frac{\beta}{2} {\bar\W_{\otimes n}}^\top\X^{-1}{\bar\W_{\otimes n}},
\end{align}
 we obtain:
 \begin{align}\label{eq:logZ}
	\mathbb{E}_\mathcal{D}{\log Z} &= \lim_{n\rightarrow 0}\frac{1}{n}\big(\mathbb{E}_\mathcal{D}{Z^n}-1\big),\nonumber\\
	\mathbb{E}_\mathcal{D}{Z^n} &= e^{\frac{n(n+1)}{2}\log(\frac{i{P}}{2\pi})+\frac{nM}{2}\log \frac{2\pi}{\beta}}\int \bigg({\prod_{a\geq b}} d\C^{ab}d\hat\C^{ab}\bigg)e^{-n{P}  S[\C,\hat\C]}.
\end{align}
The reader may question the dependence of various quantities in $S$ on $P$, since we are taking a $P\to\infty$ limit. This is because we want to keep our treatment general. Depending on the kernel and data distribution, there are other quantities here that can scale with $P$. Specific examples will be given.

\subsection{Replica Symmetry and Saddle Point Equations}
To proceed with the saddle point integration, we further assume replica symmetry relying on the convexity of the problem:
 \begin{align}
	&C^0= \C^{aa},\,\, &&\hat C^0 = \hat \C^{aa}, \nonumber \\
	&C= \C^{a\neq b},\,\, &&\hat C = \hat \C^{a\neq b}.
\end{align}
Therefore, we have $\C = (C_0-C)\I+C\1\1^\top$ and $\hat\C = (\hat C_0-\hat C)\I+\hat C\1\1^\top$. In this case, the matrix $\X$ has the form:
 
\begin{align}
    \X = \begin{pmatrix} \X_0 & \X_1 & \X_1 & \dots & \X_1 \\
    \X_1 & \X_0 &  \X_1 & \hdots & \X_1\\
    \X_1 & \X_1 & \X_0 & \dots & \vdots\\
    \vdots & \dots&  \dots & \ddots  & \vdots\\
    \X_1 & \dots & \dots & \dots & \X_0
    \end{pmatrix} = \I_{n\times n}\otimes (\X_0-\X_1)_{M\times M} + \1_{n\times n}\otimes (\X_1)_{M\times M},
\end{align}

where:
 
\begin{align}
    \X_0 &\equiv \X^{aa} = \big(\I-\bchi\bchi^\top\big)-\frac{{2P\hat C_0}}{\beta}\bLambda\nonumber\\
    \X_1 &\equiv \X^{a\neq b} =-\frac{{P\hat C}}{\beta}\bLambda.\nonumber
\end{align}

It is straightforward to calculate the inverse of this matrix using Sherman-Morrison-Woodbury formula $(A+B)^{-1} = A^{-1}-A^{-1}BA^{-1}(I+BA^{-1})^{-1}$:
 
\begin{align}
    \X^{-1} &= \I_n \otimes (\X_0-\X_1)^{-1} - \big(\1_n\otimes (\X_0-\X_1)^{-1}\X_1(\X_0-\X_1)^{-1}\big) \big(\I_n\otimes\I_M + \1_n\otimes \X_1 (\X_0-\X_1)^{-1}\big)^{-1}\nonumber\\
    &= \I_n \otimes \Q^{-1} - \1_n\otimes \Q^{-1}\X_1\Q^{-1}+ \mathcal{O}(n)\nonumber,
\end{align}

where we defined,
 
\begin{align}
    \Q \equiv \X_0-\X_1 = \I -\frac{{P(2\hat C_0-\hat C)}}{\beta}\bLambda -\bchi\bchi^\top,
\end{align}

for shorthand notation. Hence, we get:
 
\begin{align}
    {\bar\W_{\otimes n}}^\top\X^{-1}{\bar\W_{\otimes n}} = n \bar\W^\top\Q^{-1}\bar\W + \mathcal{O}(n^2).
\end{align}

We also need to calculate the determinant of this matrix which can be done by using Gaussian elimination method to bring it into a block-triangular form. The result is:
 
\begin{align}
    \det\X &= \det(\X_0-\X_1)^{n-1}\det(\X_0 + (n-1)\X_1) = \det(\X_0-\X_1)^{n-1}\det(\X_0-\X_1 + n\X_1).
\end{align}

Taylor expanding the last term using $\det(\I+ n\C) = 1+n\Tr\C + \mathcal{O}(n^2)$, we obtain:
 
\begin{align}
    \log\det\X =& n\log\det\Q+n\Tr(\X_1\Q^{-1}) + \mathcal{O}(n^2) 
    = n\log\det\Q - n\frac{P\hat C}{\beta}\Tr\bLambda\Q^{-1}+ \mathcal{O}(n^2). 
\end{align}

Next, using the matrix determinant lemma $\det(A+uv^T) = \det(A)(1+v^TA^{-1}u)$, we obtain:
 \begin{equation}
\begin{split}
	\det(\I +\frac{\beta}{\lambda}\C) &= \big[1+\frac{\beta}{\lambda}(C_0-C)\big]^n\bigg(1+n\frac{\beta C}{\lambda+\beta(C_0-C)}\bigg),\\
	\Rightarrow \log\det(\I +\frac{\beta}{\lambda}\C) &=  n\log(1+\frac{\beta}{\lambda}(C_0-C))+n\frac{\beta C}{\lambda+\beta(C_0-C)},
\end{split}
\end{equation}
Finally, we need to simplify ${\sum_{a\geq b}}\hat \C^{ab}(\C^{ab}-\bSigma^{ab})$ under the replica symmetry up to leading order in $n$:
 \begin{equation}
	{\sum_{a\geq b}}\hat \C^{ab}(\C^{ab}-\bSigma^{ab}) = n\big(\hat C_0(C_0-\varepsilon^2)-\frac{1}{2}\hat C(C-\varepsilon^2)\big).
\end{equation}
Therefore, under replica symmetry, the function $S$ given in \eqref{eq:actions} simplifies to:
 \begin{align}\label{appendix:final_action}
	S[\C,\hat\C] =& \hat C_0(C_0-\varepsilon^2)-\frac{1}{2}\hat C(C-\varepsilon^2)+\frac{1}{2}\log(1+\frac{\beta}{\lambda}(C_0-C))+\frac{1}{2}\frac{\beta C}{\lambda+\beta(C_0-C)}\nonumber \\
	&+ \frac{1}{2P}\left(\log\det\Q - \frac{P\hat C}{\beta}\Tr\bLambda\Q^{-1}\right) - \frac{\beta}{2P} \bar\W^\top\Q^{-1}\bar\W + \frac{\beta}{2P}(\bar\W-\bxi)^\top\bar\w,
\end{align}
where we recall that $\Q = \I -\frac{{P(2\hat C_0-\hat C)}}{\beta}\bLambda-\bchi\bchi^\top$. The saddle point equations of $S$ with respect to $C_0$ and $C$ are simple:
 \begin{alignat}{2}
	\frac{\partial S}{\partial C} & = 0 &&\Rightarrow \boxed{\hat C =\ \frac{\beta^2 C}{\big(\lambda+\beta(C_0-C)\big)^2}},\nonumber\\
	\frac{\partial S}{\partial C_0} & = 0 &&\Rightarrow \boxed{\hat C_0 = \frac{1}{2}\hat C -\frac{1}{2}\frac{\beta}{\lambda+\beta(C_0-C)}}.
\end{alignat}
The equation $\partial S/\partial \hat C = 0$ yields:
 
\begin{align}
    C = &\frac{P\hat C}{\beta^2}\Tr\bLambda\Q^{-1}\bLambda \Q^{-1} + \bar\W^\top\Q^{-1}\bLambda \Q^{-1}\bar\W + \varepsilon^2,
\end{align}

and the equation $\partial S/\partial \hat C_0 = 0$ yields:
 
\begin{align}
    C_0 = C + \frac{1}{\beta}\Tr\bLambda\Q^{-1}.
\end{align}

Two commonly appearing forms are:
 \begin{equation}
\begin{split}
	\kappa \equiv \lambda + & \beta(C_0-C) = \lambda + \Tr\bLambda\Q^{-1},\\
	\frac{2\hat C_0-\hat C}{\beta} &= -\frac{1}{\lambda+\beta(C_0-C)} = -\frac{1}{\kappa}.
\end{split}
\end{equation}
Plugging second equation to the expression for $\G$, we get:
 
\begin{align}
    \Q = \I +\frac{P}{\kappa}\bLambda - \bchi\bchi^\top,
\end{align}
hence we obtain the following implicit equation:
 \begin{equation}
	\kappa = \lambda + \Tr \bLambda\bigg(\I +\frac{P}{\kappa}\bLambda- \bchi\bchi^\top\bigg)^{-1}.
\end{equation}
In terms of $\kappa$, final saddle point equations reduce to:
 \begin{alignat}{2}\label{eq:saddlePoint}
	\hat C_0^* &= \frac{1}{2}\hat C^* -\frac{1}{2}\frac{\beta}{\kappa},\nonumber\\
	\hat C^* &=\frac{\beta^2 C^*}{\kappa^2},\nonumber\\
	C_0^* &=  C^* + \frac{\kappa - \lambda}{\beta},\nonumber\\
	C^* &= C^*\frac{P}{\kappa^2}\Tr\bLambda \Q^{-1}\bLambda \Q^{-1} + \bar\W^\top\Q^{-1}\bLambda \Q^{-1}\bar\W + \varepsilon^2.
\end{alignat}
Here, $^*$ indicates the quantities that give the saddle point. Finally, solving for $C^*$ in the last equation, we obtain:
 \begin{align}
	C^* &= \frac{1}{1-\frac{P}{\kappa^2}\Tr\bLambda \Q^{-1}\bLambda \Q^{-1}}\bigg(\bar\W^\top\Q^{-1}\bLambda \Q^{-1}\bar\W + \varepsilon^2\bigg).
\end{align}
Having obtained the saddle points, we can evaluate the saddle point integral. In the limit $P\to\infty$, the dominant contribution is:
 \begin{equation}
	\mathbb{E}_\mathcal{D}{Z^n} \approx e^{-nPS[\C^*,\hat\C^*]}.
\end{equation}
Taking the $n\to 0$ limit and plugging in the saddle point solutions to the expression \eqref{appendix:final_action}, we obtain the free energy $\mathbb{E}_\mathcal{D}{\log Z} = -PS$ to be:
 
\begin{align}\label{appendix:final_logZ}
\mathbb{E}_\mathcal{D}{\log Z} &= \frac{P}{2}\frac{\kappa-\lambda}{\kappa}-\frac{P}{2}\log\frac{\kappa}{\lambda}-\frac{P}{2}\log\det\Q-\frac{\beta P}{2}\frac{\varepsilon^2}{\kappa} + \frac{\beta}{2}\bar\W^\top\Q^{-1}\bar\W - \frac{\beta}{2}(\bar\W-\bxi)^\top\bar\w,\nonumber\\
   \kappa & = \lambda + \Tr\bLambda\Q^{-1},\nonumber\\
    \Q &= \I +\frac{P}{\kappa}\bLambda - \bchi\bchi^\top,\nonumber\\
    \bar \W &= (\bar\w-\bxi)-(\bchi^\top\bar\w)\bchi.
\end{align}

\subsection{Expected Estimator and the Correlation Function}
Finally, we can calculate the RKHS weights of the expected function and its variance:
 
\begin{align}
    \sqrt{\eta_\alpha}\mathbb{E}_\mathcal{D}{w^*_\alpha} &= \lim_{\beta\to\infty}\left.\frac{1}{\beta}\frac{\partial}{\partial\xi'_\alpha}\mathbb{E}_\mathcal{D}{\log Z}\right|_{{\bxi,\bchi} = 0}\nonumber\\
    \sqrt{\eta_\alpha\eta_\beta}\mathbb{E}_\mathcal{D}[w^*_\alpha(\mathcal{D}) w^*_\beta(\mathcal{D})] &= \lim_{\beta\to\infty}\left.\frac{1}{\beta}\frac{\partial^2}{\partial\chi'_\alpha\partial\chi'_\beta}\mathbb{E}_\mathcal{D}{\log Z}\right|_{{\bxi,\bchi} = 0},
\end{align}


where the derivatives are with respect to $\xi'_\alpha = \xi_\alpha/\sqrt{\eta_\alpha}$ and $\chi'_\alpha = \chi_\alpha/\sqrt{\eta_\alpha}$, respectively. Taking derivatives for each entry of $\bxi'$, we obtain:
 
\begin{align}
    \frac{1}{\beta}\frac{\partial}{\partial\xi'_\alpha}\mathbb{E}_\mathcal{D}{\log Z} = \sqrt{\eta_\alpha}\bar w_\alpha - \frac{\kappa \sqrt{\eta_\alpha}(\bar w_\alpha-\xi_\alpha)}{P\eta_\alpha + \kappa} = \frac{P\eta_\alpha\sqrt{\eta_\alpha} \bar w_\alpha + \kappa\sqrt{\eta_\alpha} \xi_\alpha}{P\eta_\alpha + \kappa}.
\end{align}

Hence the \textit{average estimator} has the following form:
 
\begin{align}\label{eq:average_estimator}
    \mathbb{E}_\mathcal{D}{f^*(\x;P)} = \sum_\rho \frac{P\eta_\rho}{P\eta_\rho + \kappa}\bar w_\rho \psi_\rho(\x),
\end{align}

which approaches to the target function as $P\to\infty$. Note that the learned function can only express the components which span the RKHS. If the target function has out-of-RKHS components, those will never be learned.

Finally, we want to calculate the correlation function of the estimator. Given the partition function $Z = \int d\w e^{-\beta H(\w;\mathcal{D})+\beta \bxi \cdot \w + \frac{\beta}{2}\bchi^\top \w \w^\top\bchi}$, notice that the variance:
 
\begin{align}
    \frac{1}{\beta^2}\frac{\partial^2}{\partial \xi_\alpha\partial \xi_\beta}\mathbb{E}_\mathcal{D}{\log Z}= \mathbb{E}_\mathcal{D}[\braket{w_\alpha w_\beta}_{\w}-\braket{w_\alpha }_{\w}\braket{w_\beta}_{\w}] = \frac{1}{\beta}\frac{\kappa}{P\eta_\alpha+\kappa}\delta_{\alpha\beta}
\end{align}

vanishes as $\beta\to\infty$ since there is a unique solution. However, there is variance to the estimator due to averaging over different training sets which is given by:
 
\begin{align}
    \mathbb{E}_\mathcal{D}[\braket{w_\alpha w_\beta}_{\w}]-\mathbb{E}_\mathcal{D}\braket{w_\alpha }_{\w}\mathbb{E}_\mathcal{D}\braket{w_\beta}_{\w},
\end{align}

and it is finite as $\beta\to\infty$. The first term, the eigenfunction expansion coefficients of the correlation function of the estimator $\braket{f(\x)f(\x')}$, can be calculated by taking two derivatives of $\mathbb{E}_\mathcal{D}{\log Z}$ with respect to $\bchi'$. To simplify the calculation, we first redefine $\Q \equiv \I +\frac{P}{\kappa}\bLambda  -\bLambda^{1/2} \bchi'\bchi'^\top\bLambda^{1/2}$ by setting $J=0$ and introduce the notation $\partial_\alpha \equiv \frac{\partial}{\partial\chi'_\alpha}$ for notational simplicity. First, we calculate the derivatives of $\kappa$:
 
\begin{align}
    \partial_\alpha\kappa &= - \Tr\bLambda \Q^{-1}\bigg(-\frac{P}{\kappa^2}\bLambda\partial_\alpha\kappa - \bLambda^{1/2} \partial_\alpha(\bchi'\bchi'^\top)\bLambda^{1/2}\bigg)\Q^{-1}\nonumber\\
    &= \partial_\alpha\kappa\frac{P}{\kappa^2}\Tr \bLambda^2\Q^{-2} + \Tr\bLambda\Q^{-1}\bLambda^{1/2}\partial_\alpha(\bchi'\bchi'^\top)\bLambda^{1/2}\Q^{-1}\nonumber\\
    &= \partial_\alpha\kappa\frac{P}{\kappa^2}\Tr \bLambda^2\Q^{-2} + 2\sum_\rho \eta_\rho \Q^{-1}_{\rho\alpha}\sqrt{\eta_\alpha}\bigg(\sum_\sigma  \sqrt{\eta_\sigma}\chi'_\sigma  \Q^{-1}_{\sigma\rho}\bigg).
\end{align}

Hence, we find that:
 
\begin{align}
    \partial_{\bchi'} \kappa|_{\bchi' = 0} = \partial_{\bchi'}\Q|_{\bchi' = 0}  = 0.
\end{align}

This greatly simplifies the second derivative of $\kappa$:
 
\begin{align}
    \partial_\alpha\partial_\beta \kappa|_{\bchi' = 0} &= \bigg[\partial_\alpha\partial_\beta\kappa \frac{P}{\kappa^2}\Tr \bLambda^2\Q^{-2} + 2\sum_\rho \eta_\rho \Q^{-1}_{\rho\alpha}\sqrt{\eta_\alpha\eta_\beta}\Q^{-1}_{\beta\rho}\bigg]\bigg|_{\bchi' = 0}\nonumber\\
    &=(\partial_\alpha\partial_\beta \kappa|_{\bchi' = 0})\sum_\rho \frac{P\eta_\rho^2}{(P\eta_\rho+\kappa)^2} +  \frac{2\kappa^2}{P} \frac{P\eta_\alpha^2}{(P\eta_\alpha+\kappa)^2}\delta_{\alpha\beta}\nonumber\\
    &= \frac{2\kappa^2}{P}\frac{1}{1-\gamma} \frac{P\eta_\alpha^2}{(P\eta_\alpha+\kappa)^2} \delta_{\alpha\beta},
\end{align}

where $\gamma = \sum_\rho \frac{P\eta_\rho^2}{(P\eta_\rho+\kappa)^2}$ as defined before. Now we calculate the variance of the expected function:
 
\begin{align}
     \sqrt{\eta_\alpha\eta_\beta}\mathbb{E}_\mathcal{D}[w^*_\alpha(\mathcal{D}) w^*_\beta(\mathcal{D})] =& \lim_{\beta\to\infty} \frac{1}{\beta}\partial_\alpha\partial_\beta \mathbb{E}_\mathcal{D}{\log Z}\big|_{\bxi',\bchi' = 0}\nonumber\\ 
     =& \frac{P}{2}\frac{\varepsilon^2}{\kappa^2}\partial_\alpha\partial_\beta\kappa + \sqrt{\eta_\alpha\eta_\beta}\bar w_\alpha \bar w_\beta-\kappa\frac{\sqrt{\eta_\alpha\eta_\beta}\bar w_\alpha \bar w_\beta}{P\eta_\beta + \kappa}-\kappa\frac{\sqrt{\eta_\alpha\eta_\beta}\bar w_\alpha \bar w_\beta}{P\eta_\alpha + \kappa}
     \nonumber\\
    &-\frac{1}{2} \bar \W^\top \Q^{-1}\bigg(-\frac{P}{\kappa^2}\bLambda\partial_\alpha\partial_\beta\kappa - \bLambda^{1/2} \partial_\alpha\partial_\beta(\bchi'\bchi'^\top)\bLambda^{1/2}\bigg)\Q^{-1}\bar\W \nonumber\\
    =&\frac{1}{1-\gamma}\bigg(\varepsilon^2+\kappa^2\sum_\rho\frac{\eta_\rho\bar w_\rho^2}{(P\eta_\rho+\kappa)^2}\bigg) \frac{P\eta_\alpha^2}{(P\eta_\alpha+\kappa)^2} \delta_{\alpha\beta}+ \kappa^2 \frac{\sqrt{\eta_\alpha\eta_\beta}\bar w_\alpha\bar w_\beta}{(P\eta_\alpha+\kappa)(P\eta_\beta+\kappa)} \nonumber\\
    &+\frac{P\eta_\beta\sqrt{\eta_\alpha\eta_\beta}\bar w_\alpha \bar w_\beta}{P\eta_\beta +\kappa}-\kappa\frac{\sqrt{\eta_\alpha\eta_\beta}\bar w_\alpha \bar w_\beta}{P\eta_\alpha + \kappa}.
\end{align}

Now we can calculate the coefficients of covariance of the estimator:
 
\begin{align}
    \sqrt{\eta_\alpha\eta_\beta}\big(\mathbb{E}_\mathcal{D}[w^*_\alpha w^*_\beta]&-\mathbb{E}_\mathcal{D}w^*_\alpha \mathbb{E}_\mathcal{D} w^*_\beta\big)\\
    &= \frac{1}{1-\gamma}\bigg(\varepsilon^2+\kappa^2\sum_\rho\frac{\eta_\rho\bar w_\rho^2}{(P\eta_\rho+\kappa)^2}\bigg) \frac{P\eta_\alpha^2}{(P\eta_\alpha+\kappa)^2} \delta_{\alpha\beta}+ \kappa^2 \frac{\sqrt{\eta_\alpha\eta_\beta}\bar w_\alpha\bar w_\beta}{(P\eta_\alpha+\kappa)(P\eta_\beta+\kappa)} \nonumber\\
    &+\frac{(P\eta_\alpha+\kappa)P\eta_\beta\sqrt{\eta_\alpha\eta_\beta}\bar w_\alpha \bar w_\beta}{(P\eta_\alpha + \kappa)(P\eta_\beta +\kappa)}-\kappa\frac{(P\eta_\beta +\kappa)\sqrt{\eta_\alpha\eta_\beta}\bar w_\alpha \bar w_\beta }{(P\eta_\alpha+\kappa)(P\eta_\beta+\kappa)} - \frac{P^2\eta_\alpha\eta_\beta\sqrt{\eta_\alpha\eta_\beta}\bar w_\alpha \bar w_\beta}{(P\eta_\alpha+\kappa)(P\eta_\beta+\kappa)}\nonumber\\
    &= \boxed{\frac{1}{1-\gamma}\bigg(\varepsilon^2+\kappa^2\sum_\rho\frac{\eta_\rho\bar w_\rho^2}{(P\eta_\rho+\kappa)^2}\bigg) \frac{P\eta_\alpha^2}{(P\eta_\alpha+\kappa)^2} \delta_{\alpha\beta}}.\nonumber
\end{align}

Hence the covariance of the estimator is:
 
\begin{align}
    Cov[\Braket{f^*(\x;P)f^*(\x';P)}_\mathcal{D}] = \sum_{\alpha\beta}\bigg(\mathbb{E}_\mathcal{D}[w^*_\alpha w^*_\beta]&-\mathbb{E}_\mathcal{D}w^*_\alpha \mathbb{E}_\mathcal{D} w^*_\beta\bigg)\psi_\alpha(\x)\psi_\beta(\x').
\end{align}

\subsection{Generalization Error}
Having computed the mean and covariance of the estimator, now we can calculate the average generalization error which can be decomposed as:
 
\begin{align}
    \mathbb{E}_\mathcal{D}{E_g} = \int d\x \, \tilde p(\x) \mathbb{E}_\mathcal{D}({f^*}^2(\x)) -2\int d\x\, \tilde p(\x) \mathbb{E}_\mathcal{D}{f^*}(\x) \bar f(\x) + \int d\x \, \tilde p(\x) \bar f(\x)^2,
\end{align}

where we compute the data average over a new distribution $\tilde p(\x)$. A useful quantity is the overlap matrix defined as:
\begin{align}
    \mathscr{O}_{\rho\gamma} = \int d\x\, \tilde p(\x) \phi_\rho(\x)\phi_\gamma(\x),
\end{align}
and $\gamma' =\sum_\rho  \mathscr{O}_{\rho\rho}\frac{P\eta_\rho^2}{(P\eta_\rho + \kappa)^2}$. In terms of these quantities, using the calculation above, we find:
 
\begin{align}
     &\int d\x\, \tilde p(\x) \mathbb{E}_\mathcal{D}({f^*}^2(\x)) - \int d\x\, \tilde p(\x) \mathbb{E}_\mathcal{D}{f^*(\x)}\mathbb{E}_\mathcal{D}{f^*(\x)}  = \frac{\gamma'}{1-\gamma}\bigg(\varepsilon^2+\kappa^2\sum_\rho\frac{\eta_\rho\bar w_\rho^2}{(P\eta_\rho+\kappa)^2}\bigg)\nonumber\\
     &\int d\x\, \tilde p(\x) \mathbb{E}_\mathcal{D}{f^*(\x)}\mathbb{E}_\mathcal{D}{f^*(\x)}  = \sum_{\rho,\gamma} \mathscr{O}_{\rho\gamma}\frac{P\eta_\rho^{3/2}\bar w_\rho}{P\eta_\rho+\kappa}\frac{P\eta_\gamma^{3/2}\bar w_\gamma}{P\eta_\gamma+\kappa}\nonumber\\
     &\int d\x\, \tilde p(\x)  \mathbb{E}_\mathcal{D}{f^*(\x)} \bar f(\x) = \sum_{\rho,\gamma} \mathscr{O}_{\rho\gamma}\sqrt{\eta_\rho}\bar w_\rho \frac{P\eta_\gamma^{3/2}\bar w_\gamma}{P\eta_\gamma+\kappa} \nonumber\\
     &\int d\x\, \tilde p(\x) \bar f(\x)^2 = \sum_{\rho,\gamma}  \mathscr{O}_{\rho\gamma} \sqrt{\eta_\rho}\bar w_\rho \sqrt{\eta_\gamma}\bar w_\gamma,
\end{align}

where the first line is the contribution to generalization error due to the estimator variance. Hence generalization error is:
 
\begin{align}
    \mathbb{E}_\mathcal{D}{E_g} &= \underbrace{\frac{\gamma'}{1-\gamma}\bigg(\varepsilon^2+\kappa^2\sum_\rho\frac{\bar a_\rho^2}{(P\eta_\rho+\kappa)^2}\bigg)}_{\textbf{Variance }V} + \underbrace{\kappa^2 \sum_{\rho\gamma}\mathscr{O}_{\rho\gamma}\frac{\bar a_\rho}{P\eta_\rho + \kappa}\frac{ \bar a_\gamma}{P\eta_\gamma + \kappa}}_{\textbf{Bias }B}
\end{align}

where we replaced $\bar a_\rho = \sqrt{\eta_\rho}\bar w_\rho$ which are the $L^2$ weights of the target. This is the bias-variance decomposition of generalization error in our setting where the bias term is monotonically decreasing while the variance term is solely responsible for any non-monotonicity appearing in the generalization error.

This expression also shows simply how to handle out-of-RKHS components of the target function. Assume that the kernel is band-limited, meaning $\eta_{\rho > N} = 0$ for some $N < M$. Then generalization error becomes:
 
\begin{align}
    \mathbb{E}_\mathcal{D}E_g &= \frac{\gamma'}{1-\gamma}\bigg(\tilde\varepsilon^2+\kappa^2\sum_{\rho=1}^N\frac{\bar a_\rho^2}{(P\eta_\rho+\kappa)^2}\bigg) + \kappa^2 \sum_{\rho,\gamma = 1}^N\mathscr{O}_{\rho\gamma}\frac{\bar a_\rho}{P\eta_\rho + \kappa}\frac{ \bar a_\gamma}{P\eta_\gamma + \kappa}\nonumber\\
    &+ 2\kappa \sum_{\rho=N+1}^M\sum_{\gamma = 1}^N\mathscr{O}_{\rho\gamma} \frac{\bar a_\rho \bar a_\gamma}{P\eta_\gamma + \kappa} + \sum_{\rho,\gamma=N+1}^M\mathscr{O}_{\rho\gamma} \bar a_\rho \bar a_\gamma,
\end{align}
where we define the effective noise $\tilde \varepsilon^2 = \varepsilon^2 + \sum_{\rho=N+1}^M\bar a_\rho$. Therefore, bias decomposes into three terms which correspond to three block components of the overlap matrix. The last diagonal block of the overlap matrix yields an irreducible error on the generalization error.

We are mostly interested in how out-of-distribution generalization deviates from when the training and test distributions are same. The in-distribution generalization is simply given by setting overlap matrix to identity:
\begin{align}
    E_g^{0,p(\x)} = \frac{\gamma}{1-\gamma}\bigg(\varepsilon^2+\kappa^2\sum_{\rho=1}^N\frac{\bar a_\rho^2}{(P\eta_\rho+\kappa)^2}\bigg) + \kappa^2\sum_{\rho=1}^N\frac{\bar a_\rho^2}{(P\eta_\rho+\kappa)^2},
\end{align}
where $E_g^{0,p(\x)}$ denotes the generalization error when both training and test distributions are $p(\x)$. Note that the data $\{\bar a_\rho\}$ and $\{\eta_\rho\}$ are obtained with respect to $p(\x)$ and can be replaced by $\{\tilde{\bar a}_\rho\}$ and $\{\tilde\eta_\rho\}$ if the test distribution is fixed and the training distribution is varied. We will first consider the former case with fixed training distribution and varying test distribution:
\begin{align}
    \Delta E_g \equiv E_g - E_g^{0,p(\x)} = \frac{\gamma'-\gamma}{1-\gamma}\varepsilon^2 + \kappa^2\bar\a^\top (P\bLambda+\kappa\I)^{-1} \mathscr{O}'(P\bLambda+\kappa\I)^{-1}\bar\a
\end{align}
where we defined $\mathscr{O}' = \mathscr{O}- \frac{1-\gamma'}{1-\gamma}\I$ which captures the effect of distribution mismatch on the generalization error. An example of shifted overlap matrix   $\mathscr{O}'$ has been shown in \figref{fig:opt_test} and \figref{fig:mnist_eg_optimization}.

\subsection{Symmetries of Overlap Matrix}\label{SI:symmetries_overlap}


The overlap matrix naturally arises when one considers the kernel eigenvalue problem with respect to two different input distributions:
\begin{align}
    \int d\x'\, p(\x') K(\x,\x') \phi_\rho(\x') &= \eta_\rho \phi_\rho(\x), \;\; \Rightarrow \;\; K(\x,\x') = \sum_\rho \eta_\rho\phi_\rho(\x)\phi_\rho(\x)\nonumber\\
    \int d\x'\, \tilde p(\x') K(\x,\x') \tilde\phi_\rho(\x') &= \tilde\eta_\rho \tilde\phi_\rho(\x),\;\; \Rightarrow \;\; K(\x,\x') = \sum_\rho \tilde\eta_\rho\tilde\phi_\rho(\x)\tilde\phi_\rho(\x)
\end{align}
Therefore we have two sets of orthonormal bases $\{\phi_\rho\}$ and $\{\tilde\phi_\rho\}$ with respect to distributions $p(\x)$ and $\tilde p(\x)$, both spanning $L^2(\mathcal{X})$. A square integrable function in this space can be expanded as:
\begin{align}
    f(\x) = \sum_\rho a_\rho \phi_\rho(\x) = \sum_\rho \tilde a_\rho \tilde\phi_\rho(\x),\quad \braket{\phi_\rho,\phi_\gamma}_p = \delta_{\rho\gamma},\;\;\braket{\tilde\phi_\rho, \tilde\phi_\gamma}_{\tilde p} = \delta_{\rho\gamma},
\end{align}
where we defined the inner product $\braket{f,g}_{p(\x)} = \int d\x\,p(\x) f(\x)g(\x)$. Expansion coefficients can be found in terms of each other via:
\begin{align}
    \tilde \a_\rho = \sum_{\gamma} a_\gamma \braket{\phi_\gamma,\tilde\phi_\rho}_{\tilde p} = (\tilde\A \a)_\rho, \quad \tilde\A_{\rho\gamma} \equiv \braket{\tilde\phi_\rho,\phi_\gamma}_{\tilde p}, \nonumber \\
    \a_\rho = \sum_{\gamma} \tilde a_\gamma \braket{\tilde\phi_\gamma, \phi_\rho}_{ p} = (\A \tilde\a)_\rho, \quad \A_{\rho\gamma} \equiv \braket{\phi_\rho,\tilde\phi_\gamma}_{p} .
\end{align}
This immediately implies that:
\begin{align}\label{eq:cross_overlap_identity}
    \tilde \A \A =  \A \tilde \A = \I
\end{align}
We call $\A, \tilde\A$ cross-overlap matrices and in general they do not correspond to norm-preserving transformations. Note that the $L^2$ norm of a function depends on the probability measure on the space:
\begin{align}
    \norm{f}^2_{p(\x)} = \sum_\rho \a_\rho^2 \neq \sum_\rho (\tilde\A\a)_\rho^2 = \norm{f}^2_{\tilde p(\x)}
\end{align}
Later we show that the Hilbert norm $\norm{f}_\mathcal{H}$ is independent of the probability measures if $f\in\mathcal{H}$.

\vskip 0.5em

\noindent These matrices also connect the eigenfunctions:
\begin{align}
    \bPhi = \tilde\A^\top\tilde\bPhi,\quad     \tilde\bPhi = \A^\top \bPhi
\end{align}
Using these relations, cross-overlap matrices can be related to overlap matrix $\mathscr{O}$:
\begin{align}
    \mathscr{O}_{\rho\gamma} &= \int\,d\x\, \tilde p(\x) \phi_\rho(\x)\phi_\gamma(\x) = \sum_{\gamma'}\tilde\A^\top_{\gamma\gamma'}\int\,d\x\, \tilde p(\x) \phi_\rho(\x)\tilde\phi_{\gamma'}(\x)  = (\tilde\A^\top\tilde\A)_{\rho\gamma} \nonumber \\
    \tilde {\mathscr{O}}_{\rho\gamma} &= \int\,d\x\,  p(\x) \tilde\phi_\rho(\x)\tilde\phi_\gamma(\x) = \sum_{\gamma'}\A^\top_{\gamma\gamma'}\int\,d\x\, p(\x) \tilde\phi_\rho(\x)\phi_{\gamma'}(\x)  = (\A^\top\A)_{\rho\gamma},
\end{align}
which, together with \eqref{eq:cross_overlap_identity}, have inverses $\mathscr{O}^{-1} = \A\A^\top$ and $\tilde{\mathscr{O}}^{-1} = \tilde\A\tilde\A^\top$. Furthermore, this shows that $\mathscr{O}$ and $\tilde{\mathscr{O}}$ are symmetric positive definite matrices.

Now we connect these matrices to the kernel eigenvalues. Kernel features defined by $\bPsi(\x) = \bLambda^{1/2} \bPhi(\x)$ are orhonormal with respect to the Hilbert inner product on the RKHS but their $L^2$ inner product depends on the probability measure:
\begin{align}\label{eq:feature_hilbert_norm}
    \braket{\bPsi(\x),\bPhi(\x)^\top}_\mathcal{H} &= \I,\quad \int d\x\,p(\x) \bPsi(\x)\bPsi(\x)^\top = \bLambda \nonumber\\
    \braket{\tilde\bPsi(\x),\tilde\bPhi(\x)^\top}_\mathcal{H} &= \I,\quad \int d\x\,\tilde p(\x) \tilde\bPsi(\x)\tilde\bPsi(\x)^\top = \tilde\bLambda,
\end{align}
where $\braket{.,.}_\mathcal{H}$ is defined in \eqref{eq:hilbert_inner_product}. In terms of features, kernel can be expressed as:
\begin{align}
    K(\x,\x') = \bPsi(\x)^\top\bPsi(\x') = \tilde\bPsi(\x)^\top\tilde\bPsi(\x').
\end{align}
We note that the Hilbert inner product of two functions $f,g\in\mathcal{H}$ does not depend on the measure against which the kernel is diagonalized \cite{RasmussenWilliams}. Hence for any function $f(\x) = \w^\top\bPsi(\x) = \tilde\w^\top\tilde\bPsi(\x)$ we have:
\begin{align}
    \norm{f}^2_\mathcal{H} = \w^\top\w = \tilde\w^\top\tilde\w,
\end{align}
which immediately implies that there exists an orthogonal transformation $U$ which rotates the features and the weights as follows:
\begin{align}
    \tilde\bPsi(\x) = U\bPsi(\x),\quad \tilde\w = U\w,
\end{align}
where $U^\top = U^{-1}$. Using the relations in \eqref{eq:feature_hilbert_norm}, one can obtain the relations:
\begin{align}
    \bLambda^{1/2}\mathscr{O}\bLambda^{1/2} = U^\top \tilde\bLambda U\nonumber\\
    \tilde\bLambda^{1/2}\tilde{\mathscr{O}}\tilde\bLambda^{1/2} = U \bLambda  U^\top.
\end{align}
Furthermore, one can easily find $U$ in terms of matrices $\A,\tilde\A$:
\begin{align}
    U = \bLambda^{1/2}\tilde \A^\top \tilde\bLambda^{-1/2},\quad U^\top = \tilde\bLambda^{1/2}\A^\top\bLambda^{-1/2}
\end{align}
Of course this relation requires eigenvalue matrices to be invertible otherwise one must replace inverses with pseudo-inverses.

Finally, we obtain how the two eigenvalue matrices connect to each other:
\begin{align}
    K(\x,\x') = \bPhi^\top \bLambda \bPhi = \tilde\bPhi^\top \tilde\bLambda \tilde\bPhi\;\Rightarrow\; \tilde\bLambda = \tilde \A \bLambda \tilde \A^\top,\quad \bLambda = \A \tilde\bLambda \A^\top.
\end{align}

These transformations help us to rewrite the generalization error in terms of the test eigenvalues and weights.

\subsection{Generalization Error In Terms of Test Distribution}

Using the identities in \ref{SI:symmetries_overlap}, we first start with expressing $\kappa$ in terms of the test distribution eigenvalues:
\begin{align}
    \kappa = \lambda + \kappa\Tr{\bLambda(P\bLambda + \kappa\I)^{-1}} = \lambda + \kappa\Tr{\tilde\bLambda(P\tilde\bLambda + \kappa\tilde{\mathscr{O}}^{-1})^{-1}},
\end{align}
where we used the identity $\tilde\bLambda^{1/2}\tilde{\mathscr{O}}\tilde\bLambda^{1/2} = U \bLambda  U^\top$ and the properties of trace. Similarly we have:
\begin{align}
    \gamma &= P\Tr(\bLambda^2(P\bLambda + \kappa\I)^{-2}) = P\Tr(\tilde\bLambda(P\tilde\bLambda + \kappa\tilde{\mathscr{O}}^{-1})^{-1}\tilde\bLambda(P\tilde\bLambda + \kappa\tilde{\mathscr{O}}^{-1})^{-1})\nonumber\\
    \gamma' &= P\Tr(\mathscr{O}\bLambda^2(P\bLambda + \kappa\I)^{-2}) = P\Tr(\mathscr{O}\tilde\bLambda(P\tilde\bLambda + \kappa\tilde{\mathscr{O}}^{-1})^{-1}\tilde\bLambda(P\tilde\bLambda + \kappa\tilde{\mathscr{O}}^{-1})^{-1})
\end{align}
Generalization error is given by:
\begin{align}
    E_g = \frac{\gamma'}{1-\gamma}\big(\varepsilon^2 + \kappa^2 \bar\a^\top (P\bLambda+\kappa\I)^{-2}\bar\a\big) + \kappa^2 \bar\a^\top (P\bLambda+\kappa\I)^{-1}\mathscr{O}(P\bLambda+\kappa\I)^{-1}\bar\a.
\end{align}
Using the relations $\bLambda = \A \tilde\bLambda \A^\top$, $\tilde\a = \tilde\A\a$ and $\tilde\A\tilde\A^\top = \tilde{\mathscr{O}}^{-1}$, we get:
\begin{align}
    E_g = \frac{\gamma'}{1-\gamma}\big(\varepsilon^2 &+ \kappa^2 \tilde{\bar\a}^\top (P\tilde\bLambda+\kappa\tilde{\mathscr{O}}^{-1})^{-1}\tilde{\mathscr{O}}^{-1}(P\tilde\bLambda+\kappa\tilde{\mathscr{O}}^{-1})^{-1}\tilde{\bar\a}\big)\nonumber\\
    &+ \kappa^2 \tilde{\bar\a}^\top (P\tilde\bLambda+\kappa\tilde{\mathscr{O}}^{-1})^{-1}\tilde{\mathscr{O}}^{-2}(P\tilde\bLambda+\kappa\tilde{\mathscr{O}}^{-1})^{-1}\tilde{\bar\a}.
\end{align}
As we have done before, we can compare this to in-distribution generalization error when both training and test distributions are $\tilde p(\x)$:
\begin{align}
    E_g^{0,\tilde p(\x)} = \frac{\tilde\gamma}{1-\tilde\gamma}\bigg(\varepsilon^2+\tilde\kappa^2\sum_{\rho=1}^N\frac{\tilde{\bar a}_\rho^2}{(P\tilde\eta_\rho+\tilde\kappa)^2}\bigg) + \tilde\kappa^2\sum_{\rho=1}^N\frac{\tilde{\bar a}_\rho^2}{(P\tilde\eta_\rho+\tilde\kappa)^2},
\end{align}
where $\tilde\kappa = \lambda + \tilde\kappa\Tr(\tilde\bLambda(P\tilde\bLambda + \tilde\kappa\I)^{-1})$ and $\tilde\gamma = P\Tr(\tilde\bLambda^2(P\tilde\bLambda + \tilde\kappa\I)^{-2})$.

\section{Further Analysis for Real Data Applications and Additional Experiments}\label{SI:further_real_data}

The form for OOD generalization with label noise $\varepsilon^2=0$ given by \eqref{eq:main_gen_error} has a simple form in terms of the overlap matrix:
\begin{align}\label{eq:SI:delta_eg}
\Delta E_g &= \v(P)^\top \mathscr{O}'\v(P)\nonumber\\
\mathscr{O}' &= \mathscr{O} - \I + \frac{\Tr \,(\mathscr{O}-\I)\M(P)}{1-\Tr \M(P)}\I\nonumber\\
\mathscr{O}_{\rho\gamma} &= \int d\x\, \tilde p(\x) \phi_\rho(\x)\phi_\gamma(\x),
\end{align}
where $\v$ and $\M$ are independent of the test distribution and only given by target weights and kernel eigenvalues with respect to training distribution. In this section, we further study how mismatched train and test distributions can improve generalization. 

\textbf{Fixed Training Distribution: } Consider minimizing $\Delta E_g$ in \eqref{eq:SI:delta_eg} with respect to the overlap matrix. For fixed training distribution, the gradient with respect to $\mathscr{O}$ will be always positive-definite since the overlap matrix appears linearly in $\Delta E_g$ and is the only test distribution related quantity. This implies that $\Delta E_g$ does not have a non-trivial extremum except for the trivial $\Delta E_g = - E_g^{0, p(\x)}$ which is possible only if $\mathscr{O} = 0$. This implies that OOD generalization can be reduced to zero and often never happens except for very special cases where all eigenfunctions become zero at some point $\x^*$  ($\phi_\rho(\x^*) = 0$) and the test distribution sharply centers around the same point $\x^*$ ($\tilde p(\x) = \delta(\x-\x^*)$). 

\begin{figure}
    \centering
    \includegraphics[width=.8\linewidth]{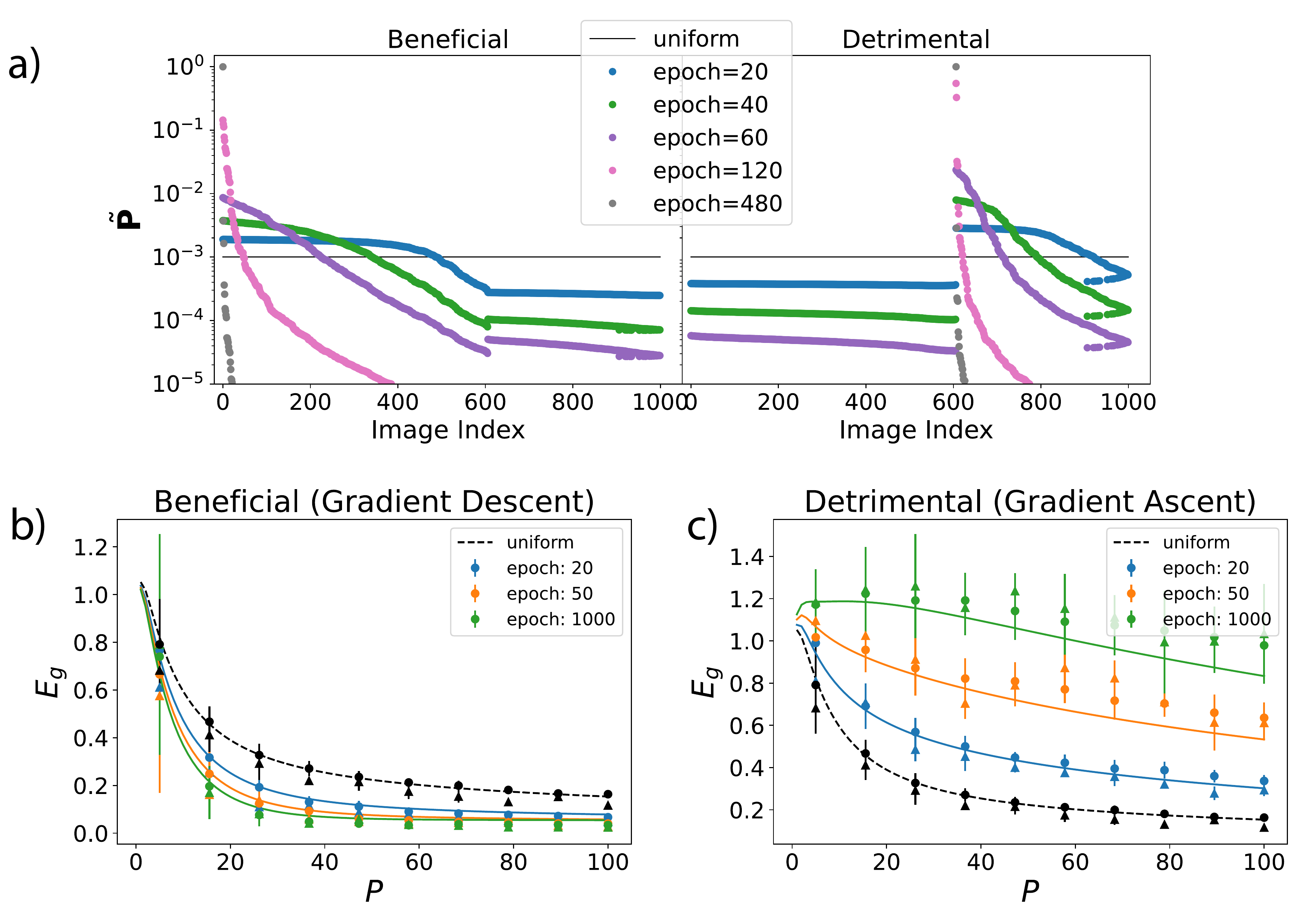}
    \caption{Optimal test distributions for fixed uniform training distribution. a) As the number of epochs increase for gradient descent/ascent, the test distribution concentrates more and more around a particular sample. We also see that the images which have high probability mass in beneficial test distribution are assigned low probability mass during gradient ascent. b) The generalization error curves for a trained neural network and kernel regression with the corresponding kernel. Both experiments match perfectly with theory (dashed lines). We see that beneficial test distributions yield better generalization than the in-distribution generalization. c) Same experiment with detrimental test distributions. As expected the generalization is worse than in-distribution case. The errorbars show standard deviation for 50 trials.}
    \label{fig:SI:fixed_train}
\end{figure}

However, a more likely scenario where $\Delta E_g$ is minimized is picking a test distribution which sharply centers around a point $\x^*$ which yields the smallest error:
\begin{align}
    \x^* = \min_{x \in \mathcal{D}} (f^*(\x) - \bar f(\x))^2.
\end{align}
Then keep testing on this single point (in the sense that $\tilde p(\x) = \delta(\x-\x^*)$) should yield the best OOD generalization hence minimizing $\Delta E_g$. This has been noted in \cite{gonzalez2015mismatched} and also what we observe in our experiments when we run gradient descent with respect to test distribution on $E_g$ for fixed training distribution. In \figref{fig:SI:fixed_train}, we show an experiment where we fix the training distribution to be uniform and obtain beneficial/detrimental test distributions by running gradient descent/ascent on $E_g$ for different epochs. In \figref{fig:SI:fixed_train} (a), we see that when number of epochs is small, the test distribution stays close to the uniform distribution. But with increasing epochs, it sharpens more and more around a particular point. Here we also show that images which have high probability mass in beneficial distribution have low probability mass in detrimental distribution. In \figref{fig:SI:fixed_train} (b,c), we show that the empirical generalization error gets smaller/larger for beneficial/detrimental test distributions and matches perfectly with our theory.

\textbf{Fixed Test Distribution: } The other way $\Delta E_g$ can be minimized is by fixing test distribution and varying training distribution. Note that, for fixed test distribution $\tilde p(\x)$, $\nabla_{\mathscr{O}|\tilde p(\x)}\Delta E_g$ is much harder to evaluate since the only way we can change overlap matrix is by changing the training distribution which in turn alters all training distribution related quantities. Hence, a priori, it is not obvious if the gradient descent procedure explained in Algorithm \ref{alg1} converges to a fixed training distribution. Next, we show that it indeed converges to a fixed training measure for MNIST experiments with NTK and we discuss its implications.

We first note that our formula for generalization error depends on the number of training set size $P$ and hence the optimized training/test measures. Then our problem is what is the optimal choice of training examples for a limited training budget $P_{\text{budget}}$ so that generalization performance is maximal. This problem intuitively makes sense; for kernel ridgeless regression it would be the best to train uniformly on all samples if we had infinite training budget since we can exactly fit all samples. Or conversely, for few shot learning (low $P_{\text{budget}}$), we would like to choose the best few training samples which yields the best generalization performance.

\begin{figure}
    \centering
    \includegraphics[width=1\linewidth]{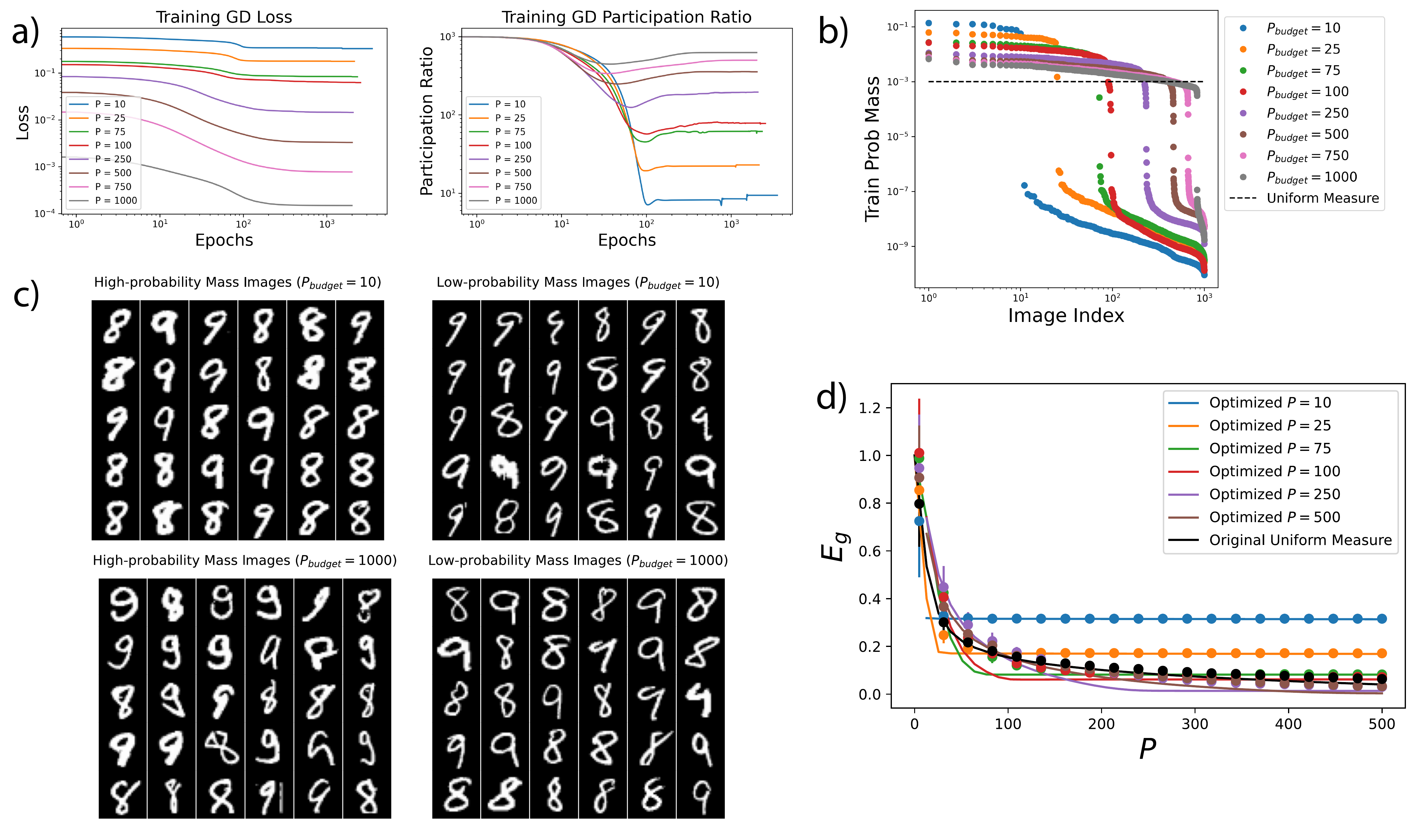}
    \caption{Optimal training distributions for fixed test distribution. a) Gradient descent on $E_g(P)$ with respect to training distribution converges for various $P_{\text{budget}}$. We also see that the participation ratio converges to $\sim P_{\text{budget}}$, implying that gradient descent finds the best $P_{\text{budget}}$ samples for fixed training budget to obtain an optimal generalization error. b) The training distributions for several $P_{\text{budget}}$'s. For larger training budgets, the training distribution approaches to the test distribution which is uniform. c) High and low probability mass images for $P_{\text{budget}} = \{10,1000\}$. For smaller budgets ($P=10$), GD finds the simplest examples to train on. For high training budgets ($P=1000$), there is no qualitative difference. d) The corresponding generalization error for each optimized training distribution compared to kernel regression experiments on NTK (dots).}
    \label{fig:SI:fixed_test}
\end{figure}

In \figref{fig:SI:fixed_test}, we show experiments where the test distributions is fixed to uniform and optimal training distributions are obtained via gradient descent on  $\Delta \bm z \propto -\nabla_{\mathscr{O}|\tilde p(\x)}\Delta E_g(P_{\text{budget}})$ for several training budgets where $\bm z$ are logits which translate to training probability masses via $\bm p = \text{softmax}(\bm z)$. In \figref{fig:SI:fixed_test}(a), we show that in fact the gradient descent converges to a fixed training distribution for all $P_{\text{budget}}$'s. Furthermore, we compute the participation ratio given by $\frac{1}{\sum_{\mu} (p^\mu)^2}$ which quantifies how uniform the probability masses are. We find that this quantity approaches roughly to $P_{\text{budget}}$ meaning that GD finds $\sim P_{\text{budget}}$ samples which are most beneficial for generalization and discards the rest. In \figref{fig:SI:fixed_test}(b), we show the training distributions for each $P_{\text{budget}}$ where each distribution is sorted from high to low probability masses. We again see that only $\sim P_{\text{budget}}$ examples have high probability mass and the rest are effectively ignored. The resulting high and low probability mass images are shown in \figref{fig:SI:fixed_test}(c). We find that for low $P_{\text{budget}}$, easier samples are preferred to train on but for high $P_{\text{budget}}$'s we do not see an apparent qualitative difference between low and high probability mass images. Finally in  \figref{fig:SI:fixed_test}(d), we test if these optimized training distributions help generalization. We find that the training distributions optimized for a certain $P_\text{budget}$ performs the best until the number of training samples hit $P_\text{budget}$ after which the generalization error stays constant.

\subsection{Adversarial Attacks during Testing}

We finally consider how our theory can be used in more practical settings such as adversarial attacks during testing. We devise a simple experiment where a kernel machine is trained on a subset of MNIST dataset with uniform training distribution and tested on the same subset but with noise added on the images. Then we identify the images which were correctly classified before adding noise but misclassified after. We create a new dataset where we add these misclassified images to the original dataset and run gradient descent/ascent on the generalization error to see if these images will be assigned low/high probability masses. We fix the training distribution to be uniform on the original images and have zero probability mass on the adversarial samples. On the other hand, the test probability masses stay as variables for the adversarial samples as well as the original samples. 

\begin{figure}
    \centering
    \includegraphics[width=1\linewidth]{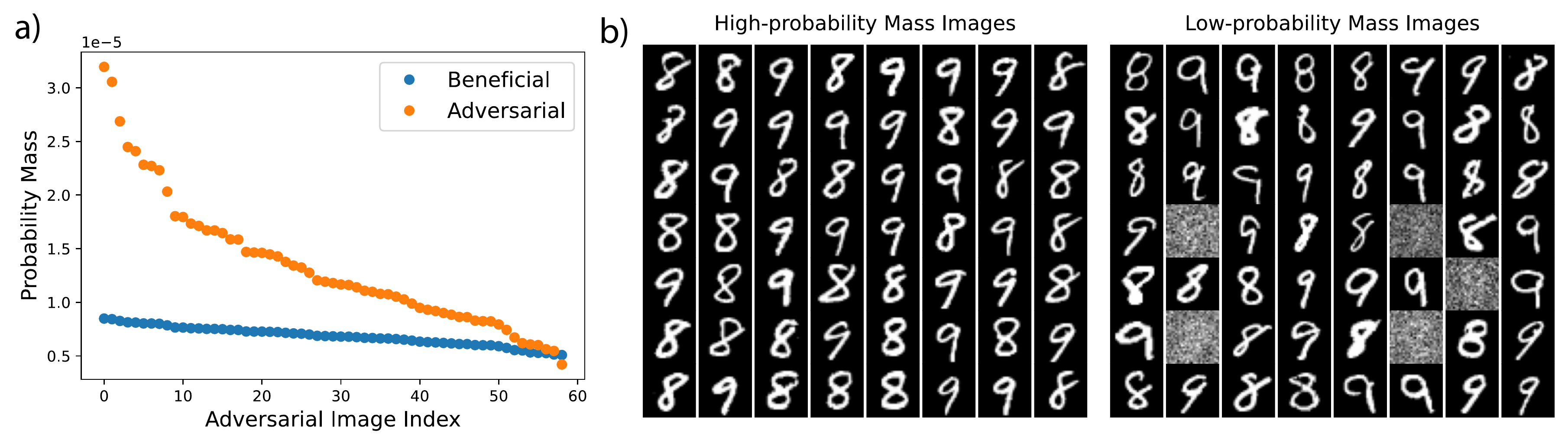}
    \caption{Adversarial samples in optimal test distribution. a) Probability masses of the adversarial samples after running gradient descent/ascent with respect to test distribution. Gradient descent/ascent places low/high probability mass to adversarial examples. b) The high and low probability mass images obtained after gradient descent. The adversarial examples get low probability mass.}
    \label{fig:SI:adversarial}
\end{figure}

In \figref{fig:SI:adversarial} (a), we show the probability masses of each adversarial MNIST images obtained after gradient descent/ascent. We find that the beneficial test distribution places considerably low probability mass to the adversarial images than the detrimental test distribution. \figref{fig:SI:adversarial} (b) shows the first few high and low probability mass images during gradient descent, where the adversarial examples are among the low probability mass images.

\section{Linear Regression}\label{SI:linear_regression}

As we discussed in the main text, it is straightforward to show that the generalization error reduces to the \eqref{eq:gen_err_gaussian} when the input distributions are Gaussian with arbitrary covariance matrices $\C$ and $\tilde\C$ for training and test distributions. 

Here, we generalize the discussion of linear regression with diagonal covariance matrices to include the out-of-RKHS scenarios. Similarly, we consider $D$-dimensional inputs and a linear target of the form $\bar f = \sum_{\rho = 1}^D \beta_\rho x_\rho$. Furthermore, we take the training and test distributions to be of the form:
\begin{align}
    \C &= \diag\, (\underbrace{\sigma^2,\dots\sigma^2}_{M_r},\underbrace{0,\dots 0}_{D-M_r})\nonumber\\
    \tilde\C &= \diag\, (\underbrace{\tilde\sigma^2,\dots\tilde\sigma^2}_{M_s},\underbrace{0,\dots 0}_{D-M_s}),
\end{align}
 where $M_r, M_s \leq D$. Finally, we allow the kernel to have less features then the ambient space so that it does not express the whole $\mathbb{R}^D$: $K(\x,\x') = \frac{1}{M}\sum_{\rho = 1}^M \x_\rho \x'_\rho$ where $M \leq D$. Therefore, we have $6$ parameters: 1) $M_r, M_s$ representing how many directions training and test distributions have non-zero variance on, 2) $N$ quantifying how many directions target depends on and $M$ how many directions kernel can represent, 3) $\sigma^2, \tilde\sigma^2$ respectively the variances of training and test distributions.

\begin{figure}[h]
    \centering
    \includegraphics[width=0.88\linewidth]{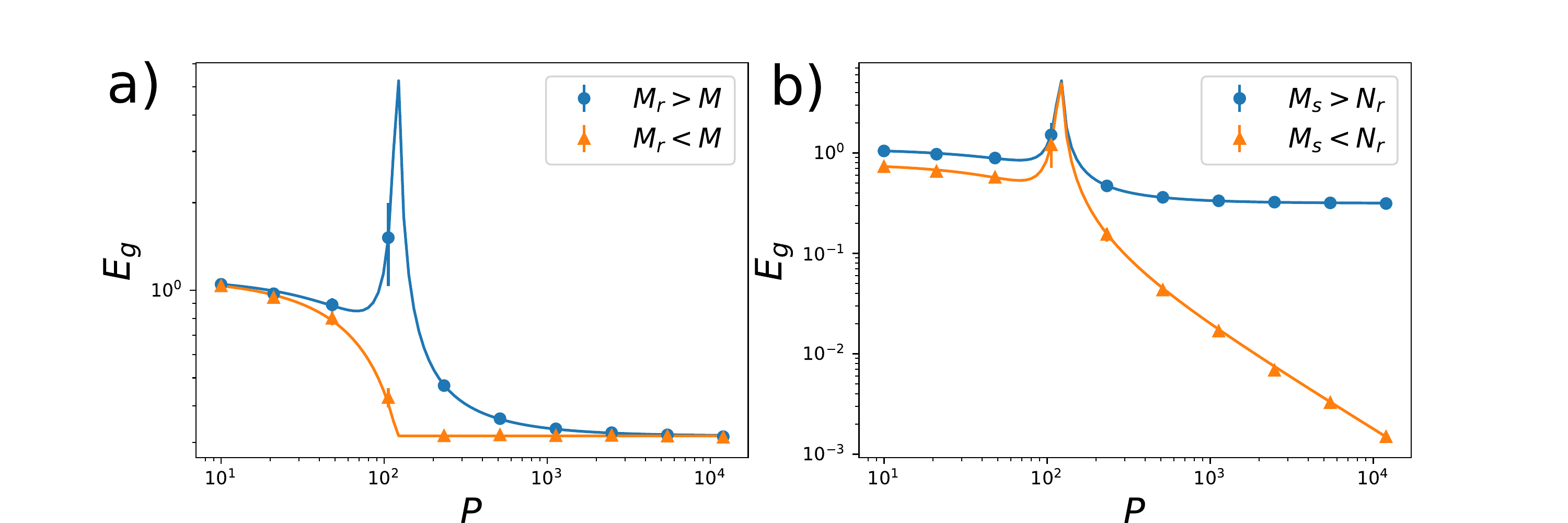}
    \caption{Effect of training and test distributions for out-of-RKHS target functions. The error bars indicate standard deviation over $30$ averages.}
    \label{fig:SI_out_of_RKHS}
\end{figure}

Then plugging the parameters of this setting in \eqref{eq:gen_err_gaussian}, the generalization error simplifies to:
\begin{align}\label{eq:general_linear_regression_case}
    \kappa' &= \frac{1}{2}\bigg[\big(1+\tilde\lambda-\alpha\big) + \sqrt{\big(1+\tilde\lambda+\alpha\big)^2-4\alpha}\bigg],\quad \tilde\lambda = \frac{\lambda}{\sigma^2 \min(1,M_r/M)},\\
    \gamma &= \frac{\alpha}{(\kappa' + \alpha)^2},\nonumber\\
    E_g &= \frac{N_{rs}}{N_r}\frac{\gamma}{1-\gamma}\bigg(\frac{\tilde\sigma^2}{\sigma^2}\varepsilon^2 + \frac{\tilde\sigma^2\kappa'^2}{(\alpha + \kappa'^2)}\sum_{\rho = 1}^{N_r}\beta_\rho^2+ \tilde\sigma^2\sum_{\rho = N_r + 1}^{M_r}\beta_\rho^2\bigg)+ \frac{\tilde\sigma^2\kappa'^2}{(\alpha + \kappa'^2)}\sum_{\rho = 1}^{N_{rs}} \beta_\rho^2 +\tilde\sigma^2\sum_{\rho = N_{rs} +1}^{M_s}\beta_\rho^2,
    \end{align}
where we defined $\alpha = P/N_r$, $N_{r} = \min\{M,M_r\}$ and $N_{rs} = \min\{M,M_r,M_s\}$. Hence, the learning rate is determined by the minimum between number of features and the number of nonzero variance directions in the training distribution. Although complicated looking, \eqref{eq:general_linear_regression_case} predicts several interesting phenomena:
\begin{enumerate}
    \item When there is an out-of-RKHS component in the target function, those components act like noise causing the effective noise given by:
    \begin{align}
        \tilde\varepsilon^2 = \tilde\sigma^2\frac{N_{rs}}{N_r}\bigg(\frac{\varepsilon^2}{\sigma^2} + \sum_{\rho = N_r + 1}^{M_r}\beta_\rho^2\bigg).
    \end{align}
    Hence, even there is no label noise in the training set, we observe a double-descent due to the effective noise.
    
    Our theory suggests that the only way avoiding the contribution to noise from out-of-RKHS components is to not train on those directions, i.e. setting $M_r \leq M$. This has been demonstrated in \figref{fig:SI_out_of_RKHS}(a).
    
    \item When the target depends on all dimensions, the irreducible error (last term) can be avoided if $M_s < N_r$. This simply means that only testing on the features which are not learned (either due to not training on those directions or the kernel not expressing them as features) causes an irreducible error in $E_g$. This has been demonstrated in \figref{fig:SI_out_of_RKHS}(b). Note that there is still double-descent since we are still training on the directions kernel does not express.
    
    \item If the irreducible error is avoided by not testing on the directions where the training distribution has zero variance or kernel does not express ($M_s \leq M, M_r$), the generalization error approaches to $0$ as $P\to\infty$ even though the target has out-of-RKHS components (See \figref{fig:SI_out_of_RKHS}(b)).
\end{enumerate}

\section{Rotation Invariant Kernels and Neural Tangent Kernel}\label{SI:ntk}

Another application of our theory is the study of rotation invariant kernels on high-dimensional input spaces. This type of kernels includes many popular kernels including Laplace kernels, radial basis function kernels and neural tangent kernel (NTK). Specifically, the kernel only depends on the inner product of two inputs: $K(\x,\x') = K(\x\cdot\x')$. In this case Mercer's decomposition takes the form:
\begin{align}
    K(\x,\x') = \sum_{k,m, n} \eta_{k,m,n} R_n(\norm{\x})R_n(\norm{\x'}) Y_{k,m}(\x)Y_{k,m}(\x')
\end{align}
where $Y_{km}$ are the hyper-spherical harmonics \cite{dai2013spherical} which only depend on the angular coordinates of $\x$ and $R_n$ depend on the norm of the input which are usually orthonormal polynomials of order $n$. If the kernel is rotation invariant, the eigenvalues only depend on the degree of the spherical harmonics $k$ but not on the azimuthal coordinates $m$ which means $N(D,k) \sim \mathcal{O}(D^k)$ times degeneracy for each $\eta_{k,m,n} = \eta_{k,n}$. Note that $\eta_{k,n} \sim \mathcal{O}(N(D,k)^{-1})$ for kernel to have finite trace. Then we define $\mathcal{O}_D(1)$ quantity $\bar\eta_{k,n} \equiv N(D,k)\eta_{k,n}$. 

Furthermore, we assume that the target function $\bar f(\x) = \sum_{k,m,n}\bar a_{k,m,n} R_n(\norm{\x}) Y_{k,m}(\x)$ has finite $L^2$ norm which implies that $\bar a^2_{k,n} \equiv \frac{1}{N(D,k)}\sum_m \bar a^2_{k,m,n}$ is finite for each mode $k$. Note 
that the overlap matrix $$\mathscr{O}_{kmn;k'm'n'} = \int d\x \, \tilde p(\x) R_n(\norm{\x})R_{n'}(\norm{\x}) Y_{k,m}(\x)Y_{k',m'}(\x)$$
is in general very difficult to calculate analytically without assuming specific probability distributions. To simplify, we consider probability distributions on hyperspheres with radius $R$ and $\tilde R$ for training and test distributions, respectively. Then considering the limit $P, D\to\infty$ while keeping $\alpha_k \equiv P/N(D,k)$ finite, we find that the different degree $k$ modes decouple over angular indices leading:
\begin{align}
    E_g &= \frac{\gamma'}{1-\gamma}\bigg(\varepsilon^2+\sum_{k'>k}\sum_{n}\bar a_{k',n}^2\bigg) + \kappa^2 \sum_{n,n'}\frac{\bar a_{k,n}}{\alpha_k\bar\eta_{k,n} + \kappa}(\mathscr{O}_{nn'}+\frac{\gamma'}{1-\gamma}\I)\frac{ \bar a_{k,n'}}{\alpha_k\bar\eta_{k,n'} + \kappa}\nonumber\\
    &+ \sum_{k'>k}\sum_{nn'}\mathscr{O}_{nn'}\bar a_{k',n} \bar a_{k',n'},
\end{align}
where $\mathscr{O}_{nn'} = \int d\x \, \tilde p(\x) R_n(\norm{\x})R_{n'}(\norm{\x})$ and  
\begin{align}
    \kappa &= \lambda  + \sum_{k'>k,n}\bar\eta_{k'n} +  \kappa\sum_n \frac{\bar\eta_{k,n}}{\alpha_k\bar\eta_{k,n}+\kappa}, \; \gamma = \alpha_k \sum_n \frac{\bar\eta^2_{k,n}}{(\alpha_k\bar\eta_{k,n}+\kappa)^2},\; \gamma' = \alpha_k \sum_n \frac{\mathscr{O}_{nn}\bar\eta^2_{k,n}}{(\alpha_k\bar\eta_{k,n}+\kappa)^2}.
\end{align}
We notice that the effective regularization becomes $\tilde\lambda \propto \lambda  + \sum_{k'>k,n}\bar\eta_{k'n}$ implying that the inductive bias of the kernel machine solely depends on the training distribution and can be altered by changing it. Furthermore, the target power for $k'>k$ acts as an effective noise and this is in fact an example of out-of-RKHS generalization: in the limit $P,D\to\infty$ for finite $\alpha_k \equiv P/N(D,k)$, the modes larger than $k$ are not in the sub-RKHS defined by polynomials of degree $k$. This has also been pointed out in \cite{canatar2020spectral}. There is also an irreducible error due to the target power for $k'>k$ which depends on both training and test distribution.

\begin{figure}
    \centering
    \subfigure[$R = \tilde R = 1$]{\includegraphics[width=0.46\linewidth]{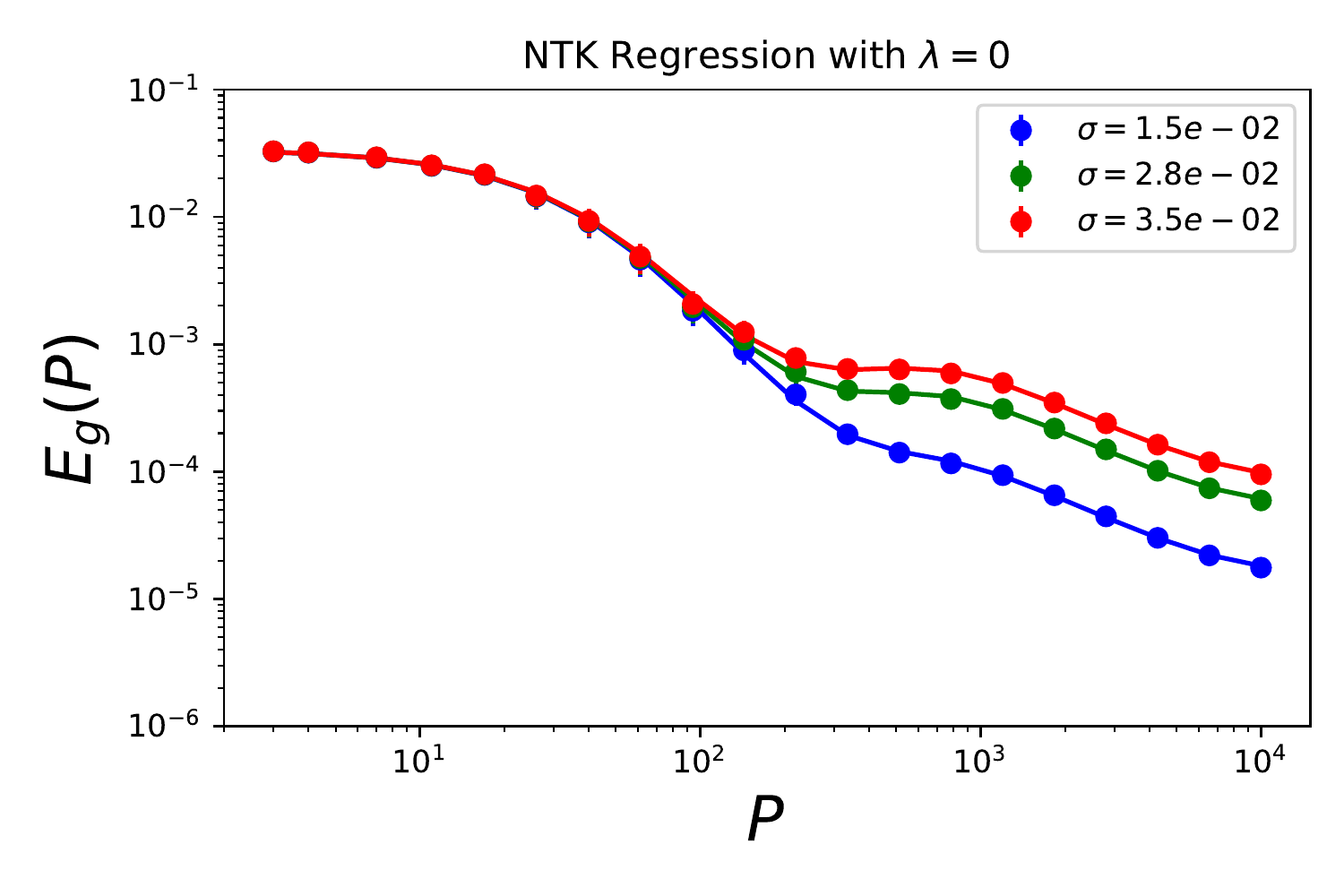}}
    \subfigure[$R = 1$ and $\tilde R = 0.5$]{\includegraphics[width=0.46\linewidth]{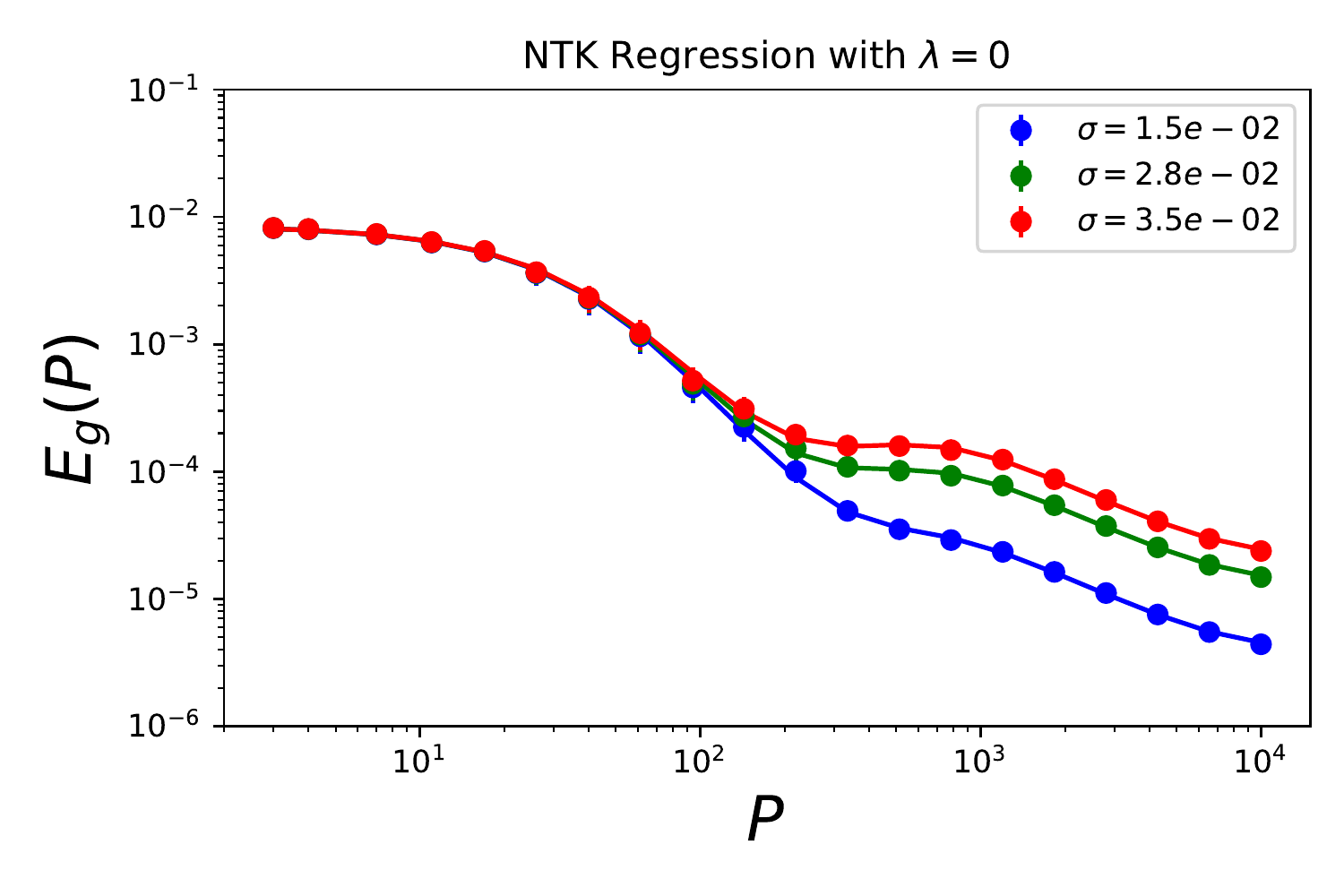}}
    \caption{2-Layer NTK Regression on random spherical data. a) The training and test distribution radii are the same $R = \tilde R = 1$. b) Training distribution radius is $R = 1$ and test distribution radius is $\tilde R = 0.5$. As predicted by theory, $E_g$ is lower in the former case.}
    \label{fig:SI_ntk_regression}
\end{figure}

To conclude this section, we finally consider ReLU NTK regression with arbitrary depth. Note that the ReLU networks are homogeneous maps with respect to the norm of inputs \cite{bietti2019inductive}:
\begin{align}\label{eq:relu_ntk_homogeneous}
    K(\x,\x') = \norm{\x}\norm{\x'}k\bigg(\frac{\x\cdot\x'}{\norm{\x}\norm{\x'}}\bigg),
\end{align}
and therefore we can drop $n$-indices in the computation above. Notice that the self-consistent equation $\kappa$ in this case can be solved exactly and the solution is very similar to the one for linear regression: $\kappa' = {\kappa/\bar\eta_k} =  \frac{1}{2}\big[(1+\tilde\lambda_k-\alpha_k) + \sqrt{(1+\tilde\lambda_k+\alpha_k\big)^2-4\alpha_k}]$ where the effective regularization is $\tilde\lambda_k = (\lambda  + \sum_{k'>k}\bar\eta_{k'})/\bar\eta_k$. In this case the learning rate is controlled by the degeneracy of mode $k$: $\alpha = P/N(D,k)$. Homogeneity of the NTK \eqref{eq:relu_ntk_homogeneous} implies that for inputs restricted to a $D$-sphere of radius $R$, the eigenvalues simply are multiplied by the norm squared: $\eta_k \to R^2\eta_k$. Therefore when an NTK regression is performed on a training set with radius $R$ and tested on a sphere with radius $\tilde R$, the overlap matrix simply becomes diagonal with components $\frac{\tilde R^2}{R^2}$ and the analysis for linear regression can be directly applied here:
\begin{align}
    E^{k}_g &= \tilde R^2\bigg(\frac{\tilde\varepsilon^2}{R^2}\frac{\alpha_k}{(\kappa' + \alpha_k)^2 - \alpha}+ \frac{\kappa'^2}{(\kappa' + \alpha_k)^2 - \alpha_k}\bigg) + \frac{\tilde R^2}{R^2}\sum_{k'>k}\bar a_k^2,
\end{align}
where we define effective noise to be $\tilde\varepsilon ^2 = \varepsilon^2 + \sum_{k'>k}\bar a_k^2$. $E^{k}_g$ is the generalization error in learning stage $k$ for large $P, D$ limit and shows that it is very similar to linear regression we studied in Section \ref{section:main_linear_regression}. At each learning stage $k$, the kernel machine learns the $k^{th}$ mode and the higher modes act as irreducible error given by $\frac{\tilde R^2}{R^2}\sum_{k'>k}\bar a_k^2$. This implies that the irreducible error is controlled by the ratio of radii of the test and target distributions. Note that for larger test radius, generalization error increases for large width neural networks. We demonstrate this in \figref{fig:SI_ntk_regression} where the inputs are randomly drawn from a sphere of radius $R$ for training set and test inputs are drawn from a sphere of radius $\tilde R$ for 2-layer NTK regression.

\section{Interpolation vs. Extrapolation}\label{SI:interpolation_extrapolation}

Another subject where understanding OOD distribution is crucial is extrapolation from training data to test data whose support lies outside the training distribution. Our theory can also explain how kernel regression generalizes in extrapolation tasks for simple models like linear regression and band-limited Fourier kernels.

To understand extrapolation in linear regression, we introduce rectangular distributions for each direction $x_\rho$ defined as:
\begin{align}\label{eq:rect_dist}
    p(\x) &= R_{\sigma_1}(x_1)\dots R_{\sigma_D}(x_D),\quad R_{\sigma_\alpha}(x) = \begin{cases} 
      \frac{1}{2\sqrt{3}\sigma_\alpha} & -\sqrt{3}\sigma_\alpha \leq x \leq \sqrt{3} \sigma_\alpha \\
      0 & \text{otherwise}
  \end{cases}\nonumber\\
  \tilde p(\x) &= R_{\tilde\sigma_1}(x_1)\dots R_{\tilde\sigma_D}(x_D).
\end{align}
Then one can take $\tilde\sigma_\alpha^2 > \sigma_\alpha^2$ to change the support of train and test distributions. Note that the Gaussian measure is an example of interpolation since the support of the data is always $\text{Supp}(p(\x)) = (-\infty,\infty)$. One can easily show that the solution to the integral eigenvalue problem for kernel is the same as when the distributions are Gaussian with diagonal covariance matrices: the kernel eigenvalues are given by $\eta_\rho = \sigma_\rho^2/D$ and eigenfunctions are $\phi_\rho(\x) = x_\rho / \sigma_\rho$ leaving the features unchanged. Then the analysis in Section \ref{section:main_linear_regression} exactly applies to the extrapolation in this scenario implying that the extrapolation and interpolation are the same when linear regression for linear tasks are concerned. For NTK, this finding has been stated in \cite{xu2021neural} that as long as all $x_\rho$ with nonzero power in target is expressed in the kernel, the kernel regression should be able to extrapolate. This intuitively makes sense since once the parameters are $\beta$ are found, the extrapolation should be trivial.

However, with nonlinear features we find that learning nonlinear functions, although possible, is much more costly when extrapolating with rectangular distributions than Gaussian distributions. Heuristically, we claim that the Gaussian distribution has always the same support and can still be thought of as interpolation no matter how much the variance is changed, while in the rectangular case supports for training and test distributions may be different and the kernel machine might not be trained on the region it is tested on. Here we demonstrate an example to explain why this is the case.

\begin{figure}[h]
    \centering
    \includegraphics[width=0.99\linewidth]{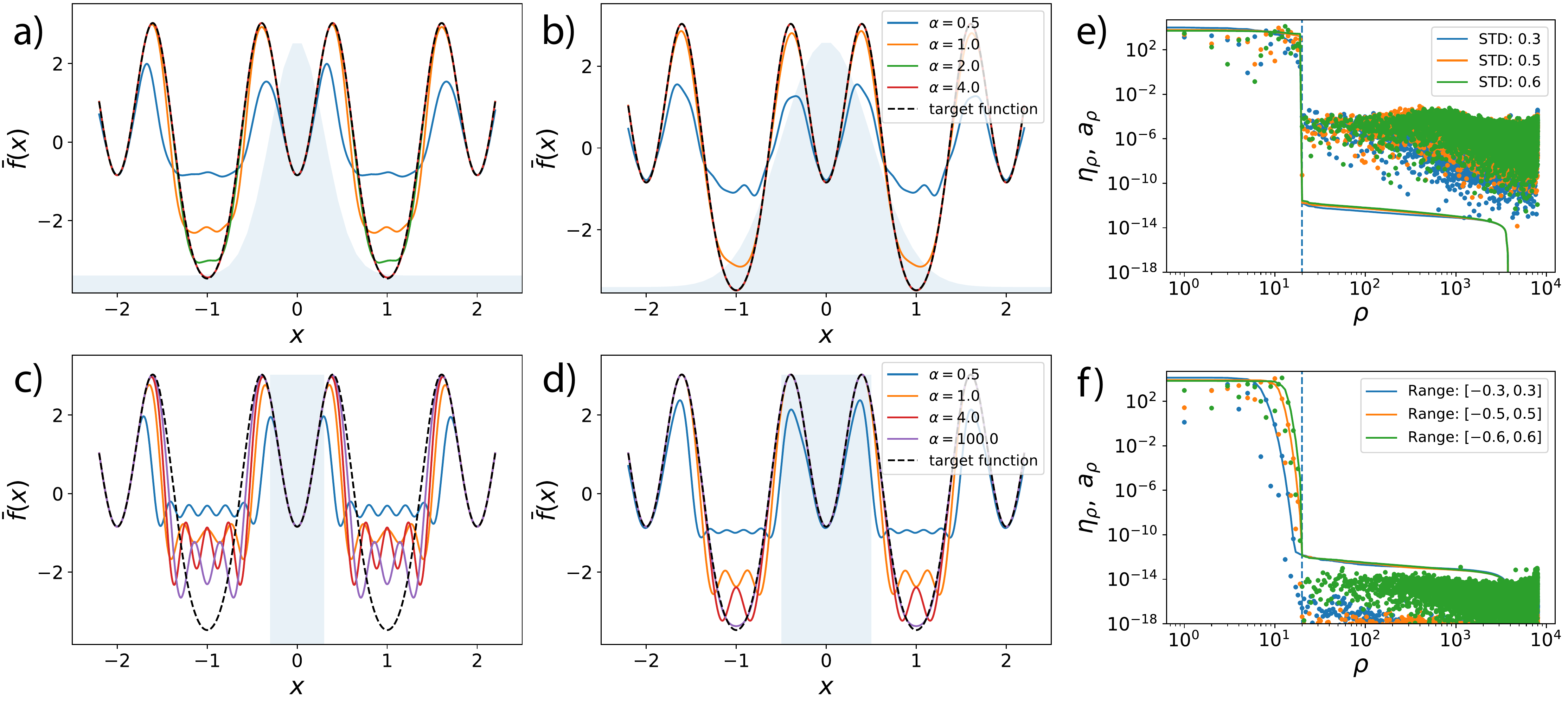}
    \caption{Interpolation with Gaussian distribution (first row) vs. Extrapolation with Rectangular distribution (second row). The estimator obtained from kernel regression is presented. a, b) Interpolation with narrower width (a) requires more training samples than wider widths (b).  c, d) In comparison, rectangular distributions work are able to interpolate well (at $\alpha = 1$) while extrapolation takes much more samples ($\alpha \sim 100$) to extrapolate. (e,f) The kernel eigenvalues on training distribution and target power are shown for Gaussian and rectangular distributions, respectively. Dashed lines indicate the number of features $N$ represented in the kernel. We observe that for varying Gaussian distribution widths, the spectrum does not change significantly while for rectangular case some eigenvalues effectively go to $0$ implying an irreducible error in the generalization.
    }
    \label{fig:extrapolation_interpolation}
\end{figure}

We consider a band-limited kernel with Fourier features in $1$D, $K(x,x') = \sum_{k=1}^N \cos k\pi(x-x') $ and a periodic target function $\bar f(x) = e^{\cos(\pi x-\theta)} + e^{\cos(\pi x+\theta)}$ centered around its mean. For the inputs, we consider centered Gaussian distributions with varying variances and rectangular distributions on the interval $x\in [-a,a]$ with varying $a$. By $\alpha = P/N$, we denote the ratio of training samples to the number of features and for certain values of $\alpha$ we compare the estimator (\eqref{eq:average_estimator}) to the target function $\bar f$. In \figref{fig:extrapolation_interpolation}(a,b), we find that the estimator perfectly matches the target when $\alpha = 4$ for both narrow and wide Gaussian training distributions. On the contrary, when the rectangular distributions are used for training in \figref{fig:extrapolation_interpolation}(c,d), we find that interpolation is achieved as soon as $\alpha = 1$ while extrapolation requires much more samples $\alpha = 250$. We attribute this behavior to the observation that some eigenvalues in the rectangular distribution case effectively goes to $0$ as can be seen from \figref{fig:extrapolation_interpolation}(f) while for Gaussian distribution they stay large \figref{fig:extrapolation_interpolation}(e). This means that as the range of the rectangular distribution gets smaller, more modes in the target function become out-of-RKHS leading to an irreducible error.

\section{Numerical Methods}\label{SI:numerical_methods}

We performed our experiments using JAX software \cite{jax2018github} and NeuralTangents package \cite{neuraltangents2020} on Google Colaboratory environment \cite{bisong2019google}. Specifically, the automatic differentiation capabilities of JAX helped us optimize over training and test distributions on MNIST digits \cite{lecun2010mnist}. All code used to perform experiments and generate figures can be accessed at \href{https://github.com/Pehlevan-Group/kernel-ood-generalization}{https://github.com/Pehlevan-Group/kernel-ood-generalization}.

\end{document}